\documentclass[preprint,12pt]{elsarticle}

\usepackage{amssymb}
\usepackage{amsmath}
\usepackage{hyperref}
\usepackage{subcaption}
\usepackage{graphicx}
\usepackage{float}
\usepackage{wrapfig}
\usepackage{comment}
\usepackage{color}
\usepackage{tabularx}
\usepackage{array}
\usepackage[table]{xcolor} % for shading
\journal{Signal Processing}
\begin{document}
\begin{frontmatter}


\title{Federated Learning: A Stochastic Approximation Approach
\tnoteref{label1}}
\tnotetext[label1]{This research did not receive any specific grant from funding agencies in the public, commercial, or not-for-profit sectors.}
\author{Srihari P V}
\ead{ee21s086@smail.iitm.ac.in}
\author{Anik Kumar Paul}
\ead{anikpaul42@gmail.com}
\author{Bharath Bhikkaji}
\ead{bharath@ee.iitm.ac.in}
\affiliation{organization={Department of Electrical Engineering, IIT Madras},
             addressline={Sardar Patel Road},
             city={Chennai},
             postcode={600036},
             state={Tamil Nadu},
             country={India}}

%% Abstract
\begin{abstract}
\allowdisplaybreaks{
 This paper considers the Federated learning (FL) in a stochastic approximation (SA) framework.  Here, each client $i$ trains a local model using its dataset $\mathcal{D}^{(i)}$ and periodically transmits the model parameters $w^{(i)}_n$ to a central server, where they are aggregated into a global model parameter  $\bar{w}_n$ and sent back. The clients continue their training by  re-initializing  their local models with the global model parameters.

%A central challenge in FL lies in understanding how heterogeneous local dynamics influence global convergence. 
Prior works  typically assumed constant (and often identical) step sizes (learning rates)  across clients for model training.  As a consequence  the aggregated model converges only in expectation. %These arguments implicitly rely on decaying learning rates for convergence, yet provide no explicit characterization of decay rates. As a result, the impact of heterogeneous or tapering step-size schedules remains unclear.
In this work,   client-specific tapering step sizes $a^{(i)}_n$ are used. The global model is shown to track an ODE with a forcing function equal to the weighted sum of the negative gradients of the individual clients. The  weights  being the limiting ratios $p^{(i)}=\lim_{n \to \infty} \frac{a^{(i)}_n}{a^{(1)}_n}$ of the step sizes, where $a^{(1)}_n \geq a^{(i)}_n, \forall n$. Unlike the constant step sizes, the convergence here is with probability one.
 
In this framework,  the clients with the larger $p^{(i)}$ exert a greater influence on the global model than those with smaller $p^{(i)}$, which can be used to favor clients that have rare and uncommon data. Numerical experiments were conducted to validate the convergence and demonstrate  the choice of step-sizes for regulating  the influence of the clients.
}
\end{abstract}


%% Keywords
\end{frontmatter}



\section{Introduction}
\label{Introduction}
\allowdisplaybreaks{

Federated learning (FL) is a framework that enables multiple users or clients to collectively train machine learning models without sharing their private data \cite{ McMahan2016CommunicationEfficientLO}.
\begin{wrapfigure}{R}{7.25cm}
\caption{Federated Learning setup with $L$ clients.}\label{figure 1}
\includegraphics[width=7.25cm]{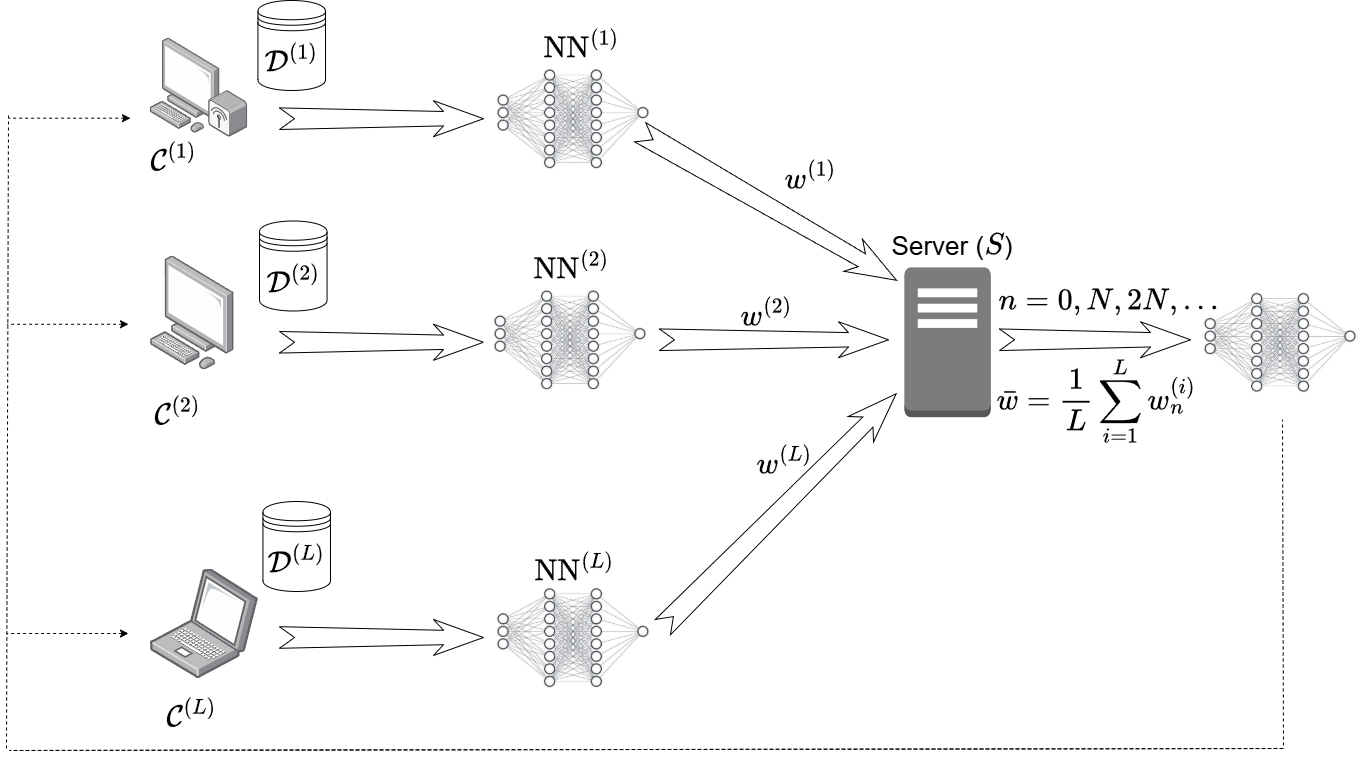}
\end{wrapfigure} 
FL has found applications in healthcare for tasks such as brain MRI analysis and tumor detection from medical images \cite{rieke_future_2020}.~It has also been used to analyze Electronic Health Records (EHR) to identify disease patterns, predict hospitalization risks, while maintaining patient data privacy \cite{rieke_future_2020}.~In the banking sector, FL is employed to predict customers loan repayment ability or credit scores without compromising their data privacy \cite{Shingi2020AFL}.

The FL framework considered here comprises of $L$ clients denoted by $\{\mathcal{C}^{(i)}\}^L_{i=1}$ and a central server $\mathcal{S}$.~Each client $\mathcal{C}^{(i)}$ has its own dataset $\mathcal{D}^{(i)}$, which is assumed to be independent of the other clients  dataset  $\mathcal{D}^{(j)}$, and is drawn from a specific distribution $\xi^{(i)}$.~The client $\mathcal{C}^{(i)}$ trains its Neural Network ${\rm NN}^{(i)}$ using the data set $\mathcal{D}^{(i)}$.~The ${\rm NN}$ architecture of the clients are assumed to be identical.~ %{\it i.e.,} the number of layers, number of neurons per layer and the activation functions in each layer of the NN are assumed to be the same. 
The weight update mechanism for these NNs are given  by 
\begin{align}
w^{(i)}_{n+1} = 
\begin{cases}
w^{(i)}_n - a^{(i)}_n \nabla_{w} E_{\xi^{(i)}}\left\{ f(w^{(i)}_n, \xi^{(i)}) \right\}, & \text{if } n \neq 0, N, 2N, 3N, \ldots \\
\frac{1}{L} \sum_{i=1}^L w^{(i)}_n, & \text{if } n = 0, N, 2N, \ldots
\end{cases}
\label{Eqn1}
\end{align}
where $w^{(i)}_n$ is the weight of the neural network for client $i$ at time step or instant $n$, $a^{(i)}_n > 0 $ is the step size for client $i$, $f$ is a differentiable cost function, $\xi^{(i)}$ the data distribution for client $i$ and $N \geq 2$ is an integer. 

Equation \eqref{Eqn1} implies that each client independently trains its NN and updates the weights at all instants $n$, except  when  $n$ is a multiple of $N$.~At these instants  the weights are sent to a common  server that aggregates the weights of all the clients.~The individual client NNs weights are then reinitialized to this aggregated weight and iterations of the clients continue until the next multiple of $N$.~This process is repeated indefinitely, Refer to Figure~\ref{figure 1} for an illustration.~The above framework applies to most literature on FL. 

Determining or computing $\nabla_{w}E_{\xi^{(i)}}\Big\{f(w^{(i)}_n, \xi^{(i)})\Big\} $ in \eqref{Eqn1} can be impractical.~A feasible alternative is to replace the Expectation term by the average 
\begin{align}
S_m^{(i)} = \frac{1}{m} \sum^m_{k=1} \nabla_w f(w^{(i)}_n,\xi^{(i)}_k),  i = 1,2, \ldots L.
\label{Eqn2}
\end{align}
If $m=1$, the iterates in \eqref{Eqn1} reduce to the stochastic gradient updates.~When $m \ > \ 1$ but significantly smaller than size of the dataset, the iterates are referred to as the mini-batch gradient updates.

Note that, replacing the expectation in \eqref{Eqn1} by \eqref{Eqn2} leads to
\begin{align}
w^{(i)}_{n+1} = 
\begin{cases}
w^{(i)}_n + a^{(i)}_n h^{(i)}(w^{(i)}_n) + a^{(i)}_n M^{(i)}_{n+1}, & \text{if } n \neq 0, N, 2N, 3N, \ldots \\
\frac{1}{L} \sum_{i=1}^L w^{(i)}_n, & \text{if } n = 0, N, 2N, \ldots,
\end{cases}
\label{Eqn3}
\end{align}
where
\begin{align}
h^{(i)}(w) &\triangleq -\nabla_w E_{\xi^{(i)}} \left\{ f(w, \xi^{(i)}) \right\}, \label{Eqn4} \\
M^{(i)}_{n+1} &\triangleq -\left( S_m^{(i)} - \nabla_w E_{\xi^{(i)}} \left\{ f(w^{(i)}_n, \xi^{(i)}) \right\} \right). \label{Eqn5}
\end{align}

In this paper algorithm \eqref{Eqn3} is analyzed in a stochastic approximation framework \cite{borkar2008stochastic}.~Under certain assumptions, the iterates of \eqref{Eqn3} are shown to track with probability one the solution of the ODE
\begin{align}
    \dot{w} = \frac{1}{L}\sum^{L}_{i=1} p^{(i)}h^{(i)}(w),
    \label{Eqn6}
\end{align}
where
\begin{align}
    p^{(i)} \triangleq \lim_{n\to \infty} \frac{a^{(i)}_n}{a^{(1)}_n}.
    \label{Eqn7}
\end{align}
It is assumed in \eqref{Eqn7}, without of loss of generality, that $a^{(1)}_n \geq  a^{(i)}_n$ and the limit exists. 

\subsection{Related Work}
FL was first proposed in~\cite{McMahan2016CommunicationEfficientLO} where two algorithms: Federated Stochastic Gradient Descent (FedSGD) and Federated Averaging (FedAvg) were introduced.~In FedSGD, the gradient in~\eqref{Eqn1} was replaced with a mini batch gradient~\eqref{Eqn2} and the number of local updates $N$ was set to  $2$ ($N=2$). The step size was taken to be constant $a^{(i)}_n = \eta > 0, \forall \ n \ {\rm and} \ i =1,2,\ldots,L$. The FedAvg had $N \geq 2$ and incorporated variable client participation (selecting a subset of clients for aggregation). The performance of the FedSGD and FedAvg algorithms were empirically validated on benchmark datasets. The convergence of the  FedAvg was analysed in \cite{DBLP:conf/iclr/LiHYWZ20}.

An improvement over FedAvg is FedProx~\cite{li2020federated}. Therein a proximal term $\frac{\mu}{2} \parallel \bar{w} - w^{(i)}_n \parallel^2$  was added  to \eqref{Eqn1} as a regularizer, % that is  the local client updates in \eqref{Eqn1} was replaced by
%\begin{eqnarray}
%w^{(i)}_{n+1} &=& w^{(i)}_n - a_n \nabla_{w}E_{\xi^{(i)}}\{f(w^{(i)}_n, \xi^{(i)})\} + \mu/2 \parallel w - w^{(i)}_n \parallel^2 
%\label{Eqn2a}
%\end{eqnarray}
where $\bar{w}$ denotes the last or the most recent averaged weight and $\mu >0$ a scalar.  This modification allowed the algorithm to indirectly address the client drift. % made the algorithm more amenable to theoretical analysis, and enabled robust convergence on datasets. % Nevertheless, convergence analysis performed on FedProx required a non-zero proximal term. %The proofs do not apply for FedAvg (when this term is absent). 
~FedNova  presented in~\cite{10.5555/3495724.3496362} uses normalized  gradients  in FedAvg and FedProx.%The normalization term compensated for both the number of local updates (for a given $N$) and the magnitude of the gradient. 
~In all the three algorithms it was proved that the   expected values  $\mathbb{E}\{F(w_n)\}$ converge to $F(w^*)$, where $w_n$ and $w^*$ denote the iterates and the minimizer respectively and $F(w)$ being the aggregated cost function.%It was further commented therein  that for  the parameters $w_n$ to converge to the minimizer $w^*$ decaying learning rates are necessary. }

%In the case of FedNova the convergence was proved for "arbitrarily small" values of step sizes. That is, the step sizes had to converge to zero.
~Note the server aggregate  in  \eqref{Eqn3} can be rewritten as 
\begin{align}
w_{n+1} 
&= w_n - \eta_n \frac{1}{L} \sum_{i=1}^L \left( w^{(i)}_n - w_n \right)&\triangleq  w_n + \eta_n \Delta_n, \quad n = 0, N, 2N, \ldots.
\label{fedopt}
\end{align}
It is evident from~\eqref{fedopt} that  a variety of algorithms can be generated by choosing different possible step sizes $\eta_n$ and the "pseudo-gradients" $-\Delta_n$.% The popular ones are the momentum-based ADAM and YOGI.
~In all the types of pseudo-gradients presented in~\cite{DBLP:conf/iclr/ReddiCZGRKKM21} the step sizes or learning rates were initialized to be the same for all clients. Asymptotic convergence was proved by letting both $L$ and $N$ go to infinity and the step sizes to zero.  %Under certain caveats the algorithms were shown to converge to the zeros of  \eqref{Eqn6} with $p^{(i)} =1$, {\it i. e.,} just simple average.

Recently, a stochastic approximation framework was applied to analyze FedAvg in~\cite{DOAN2023111294}.~Therein, the author examined the convergence under both constant and tapering step sizes.~In the case of tapering step sizes, it has been shown that the iterates converge to the zeros of~\eqref{Eqn6} with $p^{(i)}=1$. The current paper distinguishes itself from~\cite{DOAN2023111294} in three aspects, 1) an expression for the step size is provided here, %secondly, all clients uniformly use the same learning rates or step sizes, 
2) the clients can use heterogeneous learning rates or step sizes $a_n^{(i)} \neq a_n^{(j)}, i\neq j$ and 3) the algorithm provided is shown to converge to an ODE which gives more insight on the trajectory of the iterates than the pure convergence proof provided in~\cite{DOAN2023111294}.

\subsection{Contributions}
 The contributions of this paper are :
 \begin{enumerate}
    % \item A stochastic approximation framework has been applied to the 
     \item Federated Learning is analyzed in a stochastic approximation framework.  The aggregated iterates are shown to track the solution of an autonomous ODE with a forcing function being the negative of the weighted mean of the individual clients' gradients. The weights being asymptotic limits of the ratios of the learning rates $p^{(i)}$.
     \item The clients with the larger $p^{(i)}$ exert a greater influence on the global model than those with smaller $p^{(i)}$, which can be used to favor clients that have rare or uncommon data.
    \item In cases where the ODE is asymptotically stable, the iterates converge to an equilibrium point  which are the local minimizers.
    \item Numerical simulations are performed to validate the theory.
 \end{enumerate}
   
This paper is organized as follows. The underlying assumptions for the convergence analysis are presented in Section~\ref{Assumptions}. In Section~\ref{Consequence of Assumptions}, the consequences of these assumptions are summarized.  The convergence analysis is described in Section~\ref{ConvergenceAny} . The experimental results are presented in Section~\ref{Simulations}. The paper concludes in Section~\ref{Conclusion}.
}
\section{Assumptions} \label{Assumptions}
The convergence of the iterates~\eqref{Eqn3} to the solution of~\eqref{Eqn6} is proved under the following assumptions.
\begin{enumerate}
    \item [{\bf A1.}] $h^{(i)}(w)=-E_{\xi^{(i)}}\Big\{\nabla_w f(w,\xi^{(i)})\Big\}$ is Lipschitz,
    %\begin{align}
    $\parallel h^{(i)}(w_1) - h^{(i)}(w_2) \parallel \leq K^{(i)}_1 \parallel w_1 - w_2 \parallel$,
    %    \label{Eqn8}
    %\end{align}
    for some $K^{(i)}_1 > 0, \quad \forall i = 1,2,\ldots N$.
  %  Note this will be true if $\nabla_w f(w,\xi\}$ is Lipschitz.
    \item [{\bf A2.}] The step size $a^{(i)}_n = \frac{c^{(i)}}{n^{\delta_i}},\quad c^{(i)} > 0 \ \& \ \frac{3}{4} \ < \ \delta_i \leq 1$. Note
%\begin{align}
$\sum_n a^{(i)}_n = \infty$, and  %\label{Eqn9} \\
$\sum_n \left(a^{(i)}_n\right)^2 < \infty, \quad \forall i = 1, 2, \ldots, N.$ %\label{Eqn10}
%\end{align}

    \item [{\bf A3.}] $^*$ The iterates $w_n^{(i)}$ are bounded for all $i=1,2, \ldots L$ {\it i.e.,}
    %\begin{align}
      $\sup _n \parallel w_n^{(i)} \parallel^2 \ < \ \infty, \quad i=1,2,\ldots,L.$
        %\label{Eqn11}
    %\end{align}
%Note {\bf A3} holds if $f(w,\xi^{(i)})$ is convex  in $w$ or  when  the iterates are in the neighborhood of  local minima. A discussion and a proof of this is presented in Appendix C of the supplementary material.
    \item [{\bf A4.}] Random variables $\left\{\xi^{(i)}_k\right\}^{\infty}_{k=1}$ denote IID sequences with a distribution denoted by $\xi^{(i)}$ having zero mean and a finite variance for each $i=1,2,\ldots, L$. %In addition, it is assumed that $\forall k$,\quad$\xi^{(i)}_k,\quad i = 1,2, \dots L$ are mutually independent. 
    It is worth emphasising that $\xi^{(i)}$ need not be identical across $i=1,2,\ldots L$.
 \item [{\bf A5.}]  It is also assumed that 
    %\begin{align}
        $sup_{w} E_{\xi^{(i)}}\left\{\parallel \nabla_w f(w,\xi^{(i)}) \parallel^2 \right\} 
         < \infty,  \quad \forall i = 1,2,\ldots L.$
        %\label{Eqn12}
    %\end{align}
\end{enumerate}\textbf{}
$^*$ The proposed method assumes that the iterates $w^{(i)}_n$ are bounded.  It can be shown that for  convex cost functions this assumption is satisfied. It also holds true for the non-convex case  if the initialization is near a local minima.  Refer to   Appendix C of the Supplementary material for a proof.  Furthermore,  boundedness of iterates is also a standard assumption in Stochastic Approximation Literature, \cite{borkar2008stochastic}.

\subsection{Comparison of the assumptions}
The assumptions mentioned above are compared with that of the baseline algorithms FedAvg~\cite{DBLP:conf/iclr/LiHYWZ20}, FedProx~\cite{li2020federated} and FedNova~\cite{10.5555/3495724.3496362} in Table~\ref{tab:assumptions}.
\newcommand{\cmark}{\checkmark}
\newcommand{\xmark}{--}
\begin{table}[h!]
\centering
\scriptsize  % Even smaller font
\renewcommand{\arraystretch}{1.1}
\begin{tabular}{|l|c|c|c|c|}
\hline
\multicolumn{5}{|c|}{\textbf{Assumptions}} \\
\hline
\rowcolor{gray!15}
\textbf{Assumption} & \textbf{Proposed} & \textbf{FedAvg} & \textbf{FedProx} & \textbf{FedNova} \\
\hline
Smooth Lipschitz & \cmark & \cmark & \cmark & \cmark \\
\hline
Bounded derivatives & \cmark & \cmark & \xmark & \xmark \\
\hline
Bounded gradient dissimilarity & \xmark & \xmark & \cmark & \cmark \\
\hline
Convex & \cmark & \cmark & \cmark & \cmark \\
\hline
Non-convex & \cmark & \xmark & \xmark & \cmark \\
\hline
Bounded variance & \cmark & \cmark & \cmark & \cmark \\
\hline
\end{tabular}
\caption{Comparison of assumptions made for proving the convergence of the proposed method with that of the baseline algorithms.}
\label{tab:assumptions}
\end{table}
\section{Consequence of Assumptions} \label{Consequence of Assumptions}
{\allowdisplaybreaks
In this section some consequences of the assumptions made in Section~\ref{Assumptions} are presented.~These consequences will be used in the convergence proof presented in Section~\ref{ConvergenceAny}.   The proofs for these consequences are presented in the Appendix A of the  supplementary material.
\begin{enumerate}
    \item [{\bf C1.}] 
    %The sequence $\{ M^{(i)}_n\}$ in \eqref{Eqn5} is a martingale difference for all $i$.
    Due to assumption \textbf{A1}  the gradient and the expectation can be exchanged $      \nabla_wE_{{\xi}^{(i)}}\Big\{ f(w,\xi^{(i)})\Big\} =  E_{\xi^{(i)}}\Big\{\nabla_w f(w,\xi^{(i)})\Big\}.$
%    \begin{align}
%      \nabla_wE_{{\xi}^{(i)}}\Big\{ f(w,\xi^{(i)})\Big\} =  E_{\xi^{(i)}}\Big\{\nabla_w f(w,\xi^{(i)})\Big\}.
%      \label{Eqnc1}
%    \end{align}
%    Therefore
 %   \begin{eqnarray}
  %       M^{(i)}_{n+1} = -\bigg( S_m^{(i)}- E_{\xi^{(i)}}\Big\{\nabla_wf(w^{(i)}_n,\xi^{(i)})\Big\} \bigg).
  %  \end{eqnarray}
\item[{\bf C2.}] Let $\mathcal{F}_n$ be the sigma-algebra generated by the set 
$\{w^{(i)}_k, M^{(i)}_k \mid i \leq L,\; k \leq n\}$, i.e., $\mathcal{F}_n \triangleq 
\boldsymbol{\sigma}\Big( \big\{ w^{(i)}_k, M^{(i)}_k \,\big|\, i \leq L,\; k \leq n \big\} \Big).$
%\begin{equation}
%\mathcal{F}_n \triangleq 
%\boldsymbol{\sigma}\Big( \big\{ w^{(i)}_k, M^{(i)}_k \,\big|\, i \leq L,\; k \leq n \big\} \Big).
%\label{C2}
%\end{equation}
It holds that  $E_{\xi^{(i)}}\{M^{(i)}_{n+1} \mid \mathcal{F}_n\}=0$, hence $E_{\xi^{(i)}}\{M^{(i)}_{n+1}\}=0$.
\item[{\bf C3.}] The sequence $M^{(i)}_n$ have a finite variance,  {\it i. e.,}
\begin{eqnarray}
\nonumber E_{\xi^{(i)}}\left\{ \| M^{(i)}_{n+1} \|^2 \right\} 
&=& E_{\xi^{(i)}}\Big\{ \big\| S_m^{(i)} 
- E_{\xi^{(i)}}\big\{ \nabla_w f( 
\quad w^{(i)}_n, \xi^{(i)}_{n+k+1}) \big\} \big\|^2 \Big\} 
\\ &<& \infty, \quad \forall n.
\label{M_square_int2}
\end{eqnarray}
%Note, in above equations are due to $E_\xi^{(i)}\{\nabla_wf(w^{(i)}_n,\xi^{(i)})\} = -h^{(i)}(w_n)$.
\item [{\bf C4.}] Consider the sequence $R^{(i)}_{n} =\sum^{n-1}_{p=0} a_pM^{(i)}_{p+1}$.
Note, 
%\begin{eqnarray}
$E_{\xi^{(i)}}\Big\{R^{(i)}_{n+1} | \mathcal{F}_n\Big\}=R^{(i)}_{n}$.
%    \label{Martinagale2}
%\end{eqnarray}
Therefore, assuming $E_{\xi^{(i)}}\{ M^{(i)}_0 \} = 0$ , $E_{\xi^{(i)}}\Big\{R^{(i)}_{n} \Big\}=0, \forall n$. %consequence of \eqref{Exp-zero1} the following result is obtained
Furthermore, using Cauchy-Schwarz inequality it can be shown that $E_{\xi^{(i)}}\Big\{\parallel R^{(i)}_{n} \parallel^2\Big\} < \infty$.
Hence  $R^{(i)}_n$'s are zero mean square integrable martingale sequence for $i=1,2, \ldots L$.
\item [{\bf C5.}] Since
%\begin{align}
$\sum_n E_{\xi^{(i)}}\left\{ \| R^{(i)}_{n+1} - R^{(i)}_n \|^2 \right\}
= \sum_n a_n^2\, E_{\xi^{(i)}}\left\{ \| M^{(i)}_{n+1} \|^2 \right\} 
< \infty.$
%\label{martingale_bounded}
%\end{align}
by Martinagale convergence theorem \cite{williams1991probability}, the sequence $\lim_{n\to \infty}R^{(i)}_{n} = R^{(i)}_\infty$ for all $i$  almost surely, where $R^{(i)}_\infty$ is also square integrable.
 \item [{\bf C6.}]  Due to {\bf C5}, the following two must hold
\begin{align}
    \lim_{n\to \infty} \parallel a_n M_n \parallel &= 0,
    \\  \lim_{n \to \infty} \parallel \sum^{m + N}_{n=m}  a_n M_n \parallel &= 0, \;  N \geq 0.
\end{align}
Note the Martingale convergence holds even when $a_n = \frac{C}{n^p}$, where $p=\frac{1}{2} + \epsilon, \epsilon \ > \  0$. Therefore $\parallel M^{(i)}_n \parallel$ can grow   at a maximum rate of $O(n^\frac{1}{2})$. Hence 
\begin{align}
 \lim_{n \to \infty} a_{n+N}  \parallel \sum^{n+ N}_{m=n} a_m M_m \parallel &= 0, \quad {\rm and}
\\ \lim_{r \to \infty} \sum^{\infty}_{n=r} a_{n+N}  \parallel \sum^{n+ N}_{m=n} a_m M_m \parallel &= 0,
\label{EqnMart}
\end{align}
for all $a_n = \frac{C}{n^p}, \quad p=\frac{3}{4} + \epsilon$ which is assumed in {\bf A2.}

\end{enumerate}\textbf{}
}

\section{ Convergence analysis}\label{ConvergenceAny}
In this section, it will be shown that  the iterates~\eqref{Eqn3} will   asymptotically track the solution of the ODE~\eqref{Eqn6} under the assumptions made in Section~\ref{Assumptions}.  A more detailed exposition of the convergence proof can be found in the Appendix B of the Supplementary material.

%For the purpose of convergence analysis \eqref{Eqn3} can be rewritten as,
Let
\begin{align}
\bar{w}_{nN} &\triangleq  \frac{1}{L} \sum^L_{i=1} w^{(i)}_{nN}, n = 0,1,2,\ldots. 
\label{Eqn13}
\end{align}
 Using \eqref{Eqn3} 
 \begin{comment}
\begin{align}
w^{(i)}_{n+1} 
&= w^{(i)}_n - a^{(i)}_n E_{\xi^{(i)}} \left\{ \nabla_w f(w^{(i)}_n, \xi^{(i)}) \right\} \notag \\
&\quad - a^{(i)}_n \left( \frac{1}{m} \sum_{k=1}^m \nabla_w f(w^{(i)}_n, \xi^{(i)}_k) 
      - E_{\xi^{(i)}} \left\{ \nabla_w f(w^{(i)}_n, \xi^{(i)}) \right\} \right) \notag \\
&= w^{(i)}_n - a^{(i)}_n E_{\xi^{(i)}} \left\{ \nabla_w f(w^{(i)}_n, \xi^{(i)}_{n+1}) \right\} \notag \\
   &\quad - a^{(i)}_n \left( \frac{1}{m} \sum_{k=1}^m \nabla_w f(w^{(i)}_n, \xi^{(i)}_k) 
   - E_{\xi^{(i)}} \left\{ \nabla_w f(w^{(i)}_n, \xi^{(i)}) \right\} \right) \notag \\
&= w^{(i)}_n + a^{(i)}_n h^{(i)}(w^{(i)}_n) + a^{(i)}_n M^{(i)}_{n+1}.
\label{Eqn13b}
\end{align}
\end{comment}
in \eqref{Eqn13} leads to
\begin{align}
\bar{w}_{(n+1)N} 
&= \bar{w}_{nN} + \frac{1}{L} \sum_{i=1}^L \left\{ 
    \sum_{k=0}^{N-1} a^{(i)}_{nN+k} h^{(i)}(w^{(i)}_{nN+k}) 
    + \sum_{k=0}^{N-1} a^{(i)}_{nN+k} M^{(i)}_{nN+k+1} 
    \right\} \notag \\
&= \bar{w}_{nN} 
  + \frac{1}{L} \sum_{i=1}^L \sum_{k=0}^{N-1} a^{(i)}_{nN+k} h^{(i)}(w^{(i)}_{nN+k}) 
  + \frac{1}{L} \sum_{i=1}^L \sum_{k=0}^{N-1} a^{(i)}_{nN+k} M^{(i)}_{nN+k+1}.
\label{Eqn14}
\end{align}

Inductively proceeding equation~\eqref{Eqn14} leads to
\begin{align}
\bar{w}_{(n+q)N} 
&= \bar{w}_{nN} 
  + \sum_{j=0}^{q-1} \frac{1}{L} \sum_{i=1}^L \sum_{k=0}^{N-1} 
    a^{(i)}_{(n+j)N + k} \, h^{(i)}(w^{(i)}_{(n+j)N + k}) \notag \\
&\quad + \sum_{j=0}^{q-1} \frac{1}{L} \sum_{i=1}^L \sum_{k=0}^{N-1} 
    a^{(i)}_{(n+j)N + k} \, M^{(i)}_{(n+j)N + k + 1}.
\label{Eqn15}
\end{align}

The above equation can be rewritten as 
\begin{align}
\bar{w}_{(n+q)N} 
&= \bar{w}_{nN} 
  + \sum_{j=0}^{q-1} \frac{1}{L} \sum_{i=1}^L \sum_{k=0}^{N-1} 
    a^{(1)}_{(n+j)N + k}  \frac{a^{(i)}_{(n+j)N + k}}{a^{(1)}_{(n+j)N + k}} 
    h^{(i)}(w^{(i)}_{(n+j)N + k}) \notag \\
&\quad + \sum_{j=0}^{q-1} \frac{1}{L} \sum_{i=1}^L \sum_{k=0}^{N-1} 
    a^{(1)}_{(n+j)N + k}  \frac{a^{(i)}_{(n+j)N + k}}{a^{(1)}_{(n+j)N + k}} 
    M^{(i)}_{(n+j)N + k + 1} \notag \\
&\triangleq \bar{w}_{nN} 
  + \sum_{j=0}^{q-1} \frac{1}{L} \sum_{i=1}^L \sum_{k=0}^{N-1} 
    a^{(1)}_{(n+j)N + k}  p^{(i)}_{(n+j)N + k} 
    h^{(i)}(w^{(i)}_{(n+j)N + k}) \notag \\
&\quad + \sum_{j=0}^{q-1} \frac{1}{L} \sum_{i=1}^L \sum_{k=0}^{N-1} 
    a^{(1)}_{(n+j)N + k}  p^{(i)}_{(n+j)N + k} 
    M^{(i)}_{(n+j)N + k + 1},
\label{Eqn16}
\end{align}
where $q = 1,2,3,\ldots$. Due to  \eqref{Eqn7}  $p^{(i)}_n$  converges to $ p^{(i)}$.

Let $T_0 = 0$ 
\begin{align}
 T_n \triangleq  \sum^{nN}_{k=1} a^{(1)}_k, n = 1,2,3 \ldots,
\label{Eqn17}
\end{align}
$I_n = [T_n, T_{n+1}]$, $n\geq 0$. Note, $ T_n \uparrow \infty$ and $ T_{n+1} - T_n \to 0$ as $n \to \infty$.

Consider  the interpolation 
\begin{align}
\bar{w}(t) = \bar{w}_{nN}  + (\bar{w}_{(n+1)N} -\bar{w}_{nN}) \left(\frac{t-T_n}{T_{n+1} - T_n}\right), \forall t \in I_n.
\label{Eqn18}
\end{align}
Let $w^s(t)$ , $t\geq s$ denote a solution the the ODE,

\begin{align}
\dot{w}^s = \frac{1}{L}\sum^L_{i=1}p^{(i)}h^{(i)}(w^s(t));  \ w^s(s) = \bar{w}(s) \ t \geq s. 
\label{Eqn19}
\end{align}
That is,
\begin{align}
 w^s(t) = \bar{w}(s) + \frac{1}{L}\sum^{L}_{i=1}\int^t_s p^{(i)} h^{(i)}(w^s(\tau)) d\tau.
 \label{Eqn20}
 \end{align}
   Let $T > 0$ and $m, n$  be such that $T_{n+m} \in [T_n, T_n + T]$ and $[t] \triangleq \max \{ T_k \mid T_k \leq t \}$. Hence, from \eqref{Eqn20}

\begin{align}
w^{T_n}(T_{n+m}) 
&= \bar{w}_{nN} + \frac{1}{L} \sum_{i=1}^L \int_{T_n}^{T_{n+m}} p^{(i)} h^{(i)}\left(w^{T_n}(s)\right) \, ds \notag \\
&= \bar{w}_{nN} 
  + \frac{1}{L} \sum_{i=1}^L \sum_{q=0}^{m-1} \sum_{k=0}^{N-1} 
    a_{(n+q)N + k} \, p^{(i)} h^{(i)}\left(w^{T_n}(T_{n+q})\right) \notag \\
&\quad + \frac{1}{L} \sum_{i=1}^L \int_{T_n}^{T_{n+m}} 
    p^{(i)} \left( h^{(i)}\left(w^{T_n}(s)\right) 
    - h^{(i)}\left(w^{T_n}([s])\right) \right) ds.
\label{Eqn21}
\end{align}
\begin{comment}
The second summation term in~\eqref{Eqn21} is from the fact that
\begin{align}
\sum_{k=0}^{N-1} a_{(n+q)N + k} \, h^{(i)}\left(w^{T_n}(T_{n+q})\right) 
&= \int_{T_{n+q}}^{T_{n+q+1}} h^{(i)}\left(w^{T_n}(T_{n+q})\right) \, ds \notag \\
&= \int_{T_{n+q}}^{T_{n+q+1}} h^{(i)}\left(w^{T_n}([s])\right) \, ds.
\label{Eqn22}
\end{align}
\end{comment}
Due to  assumption \textbf{A1},
\begin{align}
    \parallel h^{(i)}(w) \parallel \leq \parallel h^{(i)}(0) \parallel + K_1^{(i)}\parallel w \parallel,i=1,2,\ldots, L.
    \label{Eqn23}
\end{align}
  Therefore for all $t$ such that $s\leq t \leq s+T$,
\begin{align}
\left\| w^s(t) \right\| 
&\leq \left\| \bar{w}(s) \right\| 
  + \frac{1}{L} \sum_{i=1}^L \int_s^t \left\| p^{(i)} h^{(i)}\left(w^s(\tau)\right) \right\| d\tau \notag \\
%&\leq \left\| \bar{w}(s) \right\| 
 % + \frac{1}{L} \sum_{i=1}^L \int_s^t p^{(i)} \left( \left\| h^{(i)}(0) \right\| 
%  + K_1^{(i)} \left\| w(\tau) \right\| \right) d\tau \notag \\
%&= \left\| \bar{w}(s) \right\| 
 % + \frac{1}{L} \sum_{i=1}^L \int_s^t p^{(i)} \left\| h^{(i)}(0) \right\| d\tau \notag \\
  %&\quad+ \frac{1}{L} \sum_{i=1}^L \int_s^t p^{(i)} K_1^{(i)} \left\| w(\tau) \right\| d\tau \notag \\
%&= \left\| \bar{w}(s) \right\| + T\frac{1}{L}\sum^{L}_{i=1} p^{(i)} \left\| h^{(i)}(0)\right\| 
 % + \frac{1}{L}\sum^{L}_{i=1} p^{(i)} K_1^{(i)} \int_s^t \left\| w(\tau) \right\| d\tau \notag \\
&\triangleq C_1 + K \int_s^t \left\| w(\tau) \right\| d\tau.
\label{Eqn24}
\end{align}
% where
% $\left\| h(0) \right\|= \frac{1}{L}\sum^{L}_{i=1} p^{(i)} \left\| h^{(i)}(0)\right\|$, $C_1 = \left\| \bar{w}(s)\right\|  + \left\| h(0) \right\| T $ and 
 %$\mathfrak{K_1} =\frac{1}{L}\sum^{L}_{i=1}p^{(i)}K_1^{(i)} $. 
Using Gronwalls Lemma  the solution $w^s(t)$  can be bounded by 
 \begin{align}
\left\| w^s(t) \right\| \leq C_1 e^{{K} T} \ \forall \ s \ \leq \ t \ \leq s+T.
\label{Eqn25}
\end{align}

Furthermore, 
\begin{align}
\left\| h^{(i)}\left(w^s(t)\right) \right\| 
&\leq \left\| h^{(i)}(0) \right\| + K_1 C_1 e^{K T} \notag \\
&\triangleq C_2^{(i)} 
\leq C_2, \quad \forall \ s \leq t \leq s + T.
\label{Eqn26}
\end{align}
where  $K_1 = \max_i K_1^{(i)}$ and $C_2 = \max_i C_2^{(i)}$. 

Now, if $1\leq k \leq m$ and $t\in [T_{n+k}, T_{n+k+1}] $  then
\begin{align}
\left\| w^{T_n}(t) - w^{T_n}(T_{n+k}) \right\| 
&\leq \frac{1}{L} \sum_{i=1}^L 
    \left\| \int_{T_{n+k}}^t p^{(i)} h^{(i)}\left(w^{T_n}(s)\right) ds \right\| \notag \\
%&\leq \frac{1}{L} \sum_{i=1}^L 
 %   \int_{T_{n+k}}^t p^{(i)} \left\| h^{(i)}\left(w^{T_n}(s)\right) \right\| ds \notag \\
%&\leq \frac{1}{L} \sum_{i=1}^L 
  %  \int_{T_{n+k}}^t p^{(i)} \left( \left\| h^{(i)}(0) \right\| 
 %   + K_1 C_1 e^{K T} \right) ds \notag \\
%&\leq \frac{1}{L} \sum_{i=1}^L p^{(i)} C_2 \int_{T_{n+k}}^t ds \notag \\
&\leq C \sum_{j=1}^N a^{(1)}_{(n+k)N + j}.
\label{Eqn27}
\end{align}
Consider the integral on the RHS of \eqref{Eqn21}
\begin{align}
\left\| 
\int_{T_n}^{T_{n+m}} 
\left( h^{(i)}\left(w^{T_n}(s)\right) 
- h^{(i)}\left(w^{T_n}([s])\right) \right) ds 
\right\| 
& \notag \\
%&\hspace{-10em} \leq\ 
 %K_1 \int_{T_n}^{T_{n+m}} 
 %   \left\| w^{T_n}(s) - w^{T_n}([s]) \right\| ds \notag \\
&\hspace{-10em} \leq\ 
 K_1 \sum_{k=0}^{m-1} \int_{T_{n+k}}^{T_{n+k+1}} 
    \left\| w^{T_n}(s) - w^{T_n}(T_{n+k}) \right\| ds \notag \\
%&\hspace{-10em} \leq\ 
 %K_1 C \sum_{k=0}^{m-1} \int_{T_{n+k}}^{T_{n+k+1}} 
  %  \sum_{j=1}^{N} a^{(1)}_{(n+k)N + j} \, ds \notag \\
&\hspace{-10em} \leq\ 
 K_1 C \sum_{k=0}^{m-1} \left( 
    \sum_{j=1}^{N} a^{(1)}_{(n+k)N + j} \right)^2 
 \triangleq\ 
\mathrm{K_n}.
\label{Eqn28}
\end{align}
Note, $\mathrm{K_n} \rightarrow 0$ as $n \uparrow \infty$.
In the following, it will be shown that iterates $\bar{w}_{(n+m)N}$ get  arbitrarily close to $w^{T_n}(T_{n+m})$ as $n\to \infty$. Setting $q=m$ in~\eqref{Eqn16} and subtracting it from~\eqref{Eqn20}
\begin{align}
\left\| \bar{w}_{(n+m)N} - w^{T_n}(T_{n+m}) \right\| \notag \\
&\hspace{-10em} = 
\left\| \sum_{j=0}^{m-1} \frac{1}{L} \sum_{i=1}^{L} \sum_{k=0}^{N-1} a^{(1)}_{(n+j)N+k} p^{(i)}_{(n+j)+k} h^{(i)}\left(w^{(i)}_{(n+j)N+k}\right) \right. \notag \\
&\hspace{-9em} + \sum_{j=0}^{m-1} \frac{1}{L} \sum_{i=1}^{L} \sum_{k=0}^{N-1} a^{(1)}_{(n+j)N+k} p^{(i)}_{(n+j)+k} M^{(i)}_{(n+j)N+k+1} \notag \\
&\hspace{-9em} - \frac{1}{L} \sum_{i=1}^{L} \sum_{j=0}^{m-1} \sum_{k=0}^{N-1} a^{(1)}_{(n+j)N+k} p^{(i)} h^{(i)}\left(w^{T_n}(T_{n+j})\right) \notag \\
&\hspace{-9em} \left. - \frac{1}{L} \sum_{i=1}^{L} \int_{T_n}^{T_{n+m}} p^{(i)} \left(h^{(i)}(w^{T_n}(s)) - h^{(i)}(w^{T_n}([s]))\right) ds \right\| \notag \\
&\hspace{-10em}\leq \left\| \sum_{j=0}^{m-1} \frac{1}{L} \sum_{i=1}^{L} \sum_{k=0}^{N-1} a^{(1)}_{(n+j)N+k} \left( p^{(i)}_{(n+j)+k} - p^{(i)} \right) h^{(i)}\left(w^{(i)}_{(n+j)N+k}\right) \right\| \notag \\
&\hspace{-9em} + \left\| \sum_{j=0}^{m-1} \frac{1}{L} \sum_{i=1}^{L} \sum_{k=0}^{N-1} a^{(1)}_{(n+j)N+k} p^{(i)}_{(n+j)+k} M^{(i)}_{(n+j)N+k+1} \right\| \notag \\
&\hspace{-9em} + \left\| \sum_{j=0}^{m-1} \frac{1}{L} \sum_{i=1}^{L} \sum_{k=0}^{N-1} a^{(1)}_{(n+j)N+k} p^{(i)} \left(h^{(i)}\left(w^{(i)}_{(n+j)N+k}\right) - h^{(i)}\left(w^{T_n}(T_{n+j})\right)\right) \right\| \notag \\
&\hspace{-9em} + \mathrm{K_n}
\label{Eqn29}
\end{align}

Now, Consider the  third term of~\eqref{Eqn29} 
\begin{align}
&\sum_{j=0}^{m-1} \frac{1}{L} \sum_{i=1}^L \sum_{k=0}^{N-1} a^{(1)}_{(n+j)N+k} p^{(i)} 
\left\| h^{(i)}(w^{(i)}_{(n+j)N+k}) - h^{(i)}\left(w^{T_n}(T_{n+j})\right) \right\| \notag \\
%&\quad \leq K_1 \sum_{j=0}^{m-1} \frac{1}{L} \sum_{i=1}^L \sum_{k=0}^{N-1} a^{(1)}_{(n+j)N+k} p^{(i)}
%\left\| w^{(i)}_{(n+j)N+k} - w^{T_n}(T_{n+j}) \right\| \notag \\
&\quad \leq K_1 \sum_{j=0}^{m-1} \frac{1}{L} \sum_{i=1}^L \sum_{k=0}^{N-1} a^{(1)}_{(n+j)N+k} p^{(i)}
\left\| w^{(i)}_{(n+j)N+k} - \bar{w}_{(n+j)N} \right\| \notag \\
&\qquad + K_1 \sum_{j=0}^{m-1} \frac{1}{L} \sum_{i=1}^L \sum_{k=0}^{N-1} a^{(1)}_{(n+j)N+k} p^{(i)}
\left\| \bar{w}_{(n+j)N} - w^{T_n}(T_{n+j}) \right\| \notag \\
%&\quad = K_1 \sum_{j=0}^{m-1} \frac{1}{L} \sum_{i=1}^L \sum_{k=0}^{N-1} a^{(1)}_{(n+j)N+k} p^{(i)}
%\left\| \sum_{r=0}^{k-1} a^{(i)}_{(n+j)N+r} h^{(i)}(w^{(i)}_{(n+j)N+r}) \right .\notag \\
%&\qquad+ \left .\sum_{r=0}^{k-1} a^{(i)}_{(n+j)N+r} M^{(i)}_{(n+j)N+r+1} \right\| \notag \\
%&\qquad + K_1 \sum_{j=0}^{m-1} \frac{1}{L} \sum_{i=1}^L \sum_{k=0}^{N-1} a^{(1)}_{(n+j)N+k} p^{(i)}
%\left\| \bar{w}_{(n+j)N} - w^{T_n}(T_{n+j}) \right\| \notag \\
&\quad \leq K_1 \sum_{j=0}^{m-1} \frac{1}{L} \sum_{i=1}^L \sum_{k=0}^{N-1} a^{(1)}_{(n+j)N+k} p^{(i)}
\left\| \sum_{r=0}^{k-1} a^{(i)}_{(n+j)N+r} h^{(i)}(w^{(i)}_{(n+j)N+r}) \right\| \notag \\
&\qquad + K_1 \sum_{j=0}^{m-1} \frac{1}{L} \sum_{i=1}^L \sum_{k=0}^{N-1} a^{(1)}_{(n+j)N+k} p^{(i)}
\left\| \sum_{r=0}^{k} a^{(i)}_{(n+j)N+r} M^{(i)}_{(n+j)N+r+1} \right\| \notag \\
&\qquad + K_1 \sum_{j=0}^{m-1} \frac{1}{L} \sum_{i=1}^L \sum_{k=0}^{N-1} a^{(1)}_{(n+j)N+k} p^{(i)}
\left\| \bar{w}_{(n+j)N} - w^{T_n}(T_{n+j}) \right\|.
\label{Eqn30}
\end{align}

Consider the first term of~\eqref{Eqn30}, as the iterates are bounded  there exists an $P > 0$ such that $\parallel h^{(i)}(w^{(i)}_{(n+j)N+r}) \parallel < P,\forall i$ and we get,
\begin{align}
\mathrm{P_n} 
&\triangleq K_1 \sum_{j=0}^{m-1} \frac{1}{L} \sum_{i=1}^L \sum_{k=0}^{N-1} 
a^{(1)}_{(n+j)N+k} p^{(i)} 
\left\| \sum_{r=0}^{k-1} a^{(i)}_{(n+j)N+r} h^{(i)}\left(w^{(i)}_{(n+j)N+r}\right) \right\| \nonumber \\
&\leq K_1 P \sum_{j=0}^{m-1} \frac{1}{L} \sum_{i=1}^L \sum_{k=0}^{N-1} 
a^{(1)}_{(n+j)N+k} p^{(i)} \sum_{r=0}^{k-1} a^{(i)}_{(n+j)N+r}.
\label{Eqn31}
\end{align}

Note, $\mathrm{P_{n}} \rightarrow 0, n \uparrow \infty$.
Consider the third term of~\eqref{Eqn30},
\begin{align}
& K_1 \sum_{j=0}^{m-1} \frac{1}{L} \sum_{i=1}^{L} \sum_{k=0}^{N-1} 
a^{(1)}_{(n+j)N+k} p^{(i)} 
\left\| \bar{w}_{(n+j)N} - w^{T_n}(T_{n+j}) \right\| \notag \\
&\quad \leq K_1 N \sum_{j=0}^{m-1} \frac{1}{L} \sum_{i=1}^{L} 
a^{(1)}_{(n+j)N} p^{(i)} 
\left\| \bar{w}_{(n+j)N} - w^{T_n}(T_{n+j}) \right\| \nonumber \\
&\quad \triangleq K_{Np} \sum_{j=0}^{m-1} a^{(1)}_{(n+j)N} 
\left\| \bar{w}_{(n+j)N} - w^{T_n}(T_{n+j}) \right\|.
\label{Eqn32}
\end{align}

So~\eqref{Eqn30} boils down to 
\begin{align}
&\sum_{j=0}^{m-1} \frac{1}{L} \sum_{i=1}^{L} \sum_{k=0}^{N-1} 
a^{(1)}_{(n+j)N+k} p^{(i)} 
\left\| h^{(i)}\left(w^{(i)}_{(n+j)N+k}\right) 
- h^{(i)}\left(w^{T_n}(T_{n+j})\right) \right\| \nonumber \\
&\hspace{2em} \leq \mathrm{P_{n}} 
+ K_1 \sum_{j=0}^{m-1} \frac{1}{L} \sum_{i=1}^{L} \sum_{k=0}^{N-1} 
a^{(1)}_{(n+j)N+k} p^{(i)} 
\left\| \sum_{r=0}^{k} a^{(i)}_{(n+j)N+r} M^{(i)}_{(n+j)N+r+1} \right\| \nonumber \\
&\hspace{4em} + K_{Np} \sum_{j=0}^{m-1} a^{(1)}_{(n+j)N} 
\left\| \bar{w}_{(n+j)N} - w^{T_n}(T_{n+j}) \right\| \nonumber \\
&\hspace{2em} = \mathrm{P_{n}} + \mathrm{Q_{n}} 
+ K_{Np} \sum_{j=0}^{m-1} a^{(1)}_{(n+j)N} 
\left\| \bar{w}_{(n+j)N} - w^{T_n}(T_{n+j}) \right\|.
\label{Eqn33}
\end{align}

where $\mathrm{Q_{n}}$ represents the second term that involves the Martingale sequence. Note
\begin{align}
\mathrm{Q_{n}}
&= \sum_{j=0}^{m-1} \frac{1}{L} \sum_{i=1}^{L} \sum_{k=0}^{N-1} 
a^{(1)}_{(n+j)N+k} p^{(i)} 
\left\| \sum_{r=0}^{k} a^{(i)}_{(n+j)N+r} M^{(i)}_{(n+j)N+r+1} \right\| \nonumber \\
&= \frac{1}{L} \sum_{i=1}^{L} p^{(i)} \sum_{k=0}^{N-1} \sum_{j=0}^{m-1} 
a^{(1)}_{(n+j)N+k} 
\left\| \sum_{r=0}^{k} a^{(i)}_{(n+j)N+r} M^{(i)}_{(n+j)N+r+1} \right\| \nonumber \\
&= \frac{1}{L} \sum_{i=1}^{L} p^{(i)} \sum_{k=0}^{N-1} \sum_{q=n}^{n+m-1} 
a^{(1)}_{qN+k} 
\left\| \sum_{r=0}^{k} a^{(i)}_{qN+r} M^{(i)}_{qN+r+1} \right\|.
\label{EqnQn}
\end{align}
As $N$ and $L$ are finite and due to \eqref{EqnMart}, $\mathrm{Q_{n}}$ converges to zero as $n\to\infty$.

From \eqref{Eqn29} and \eqref{Eqn33} we get,
\begin{align}
&\left\| \bar{w}_{(n+m)N} - w^{T_n}(T_{n+m}) \right\|  \notag \\
&\leq \mathrm{P_{n}} + \mathrm{Q_{n}} + \mathrm{K_{n}} 
+ K_{Np} \sum_{j=0}^{m-1} a^{(1)}_{(n+j)N} 
\left\| \bar{w}_{(n+j)N} - w^{T_n}(T_{n+j}) \right\| \nonumber \\
&\hspace{2em} 
+ \left\| \sum_{j=0}^{m-1} \frac{1}{L} \sum_{i=1}^{L} \sum_{k=0}^{N-1} 
a^{(1)}_{(n+j)N+k} \left( p^{(i)}_{(n+j)+k} - p^{(i)} \right) 
h^{(i)}\left(w^{(i)}_{(n+j)N+k}\right) \right\| \nonumber \\
&\hspace{2em} 
+ \left\| \sum_{j=0}^{q-1} \frac{1}{L} \sum_{i=1}^{L} \sum_{k=0}^{N-1} 
a^{(1)}_{(n+j)N+k} p^{(i)}_{(n+j)+k} 
M^{(i)}_{(n+j)N+k+1} \right\| \nonumber \\
&\triangleq \mathrm{P_{n}} + \mathrm{Q_{n}} + \mathrm{K_{n}}  + \mathrm{W_{n}} + \mathrm{U_{n}}
+ K_{Np} \sum_{j=0}^{m-1} a^{(1)}_{(n+j)N} 
\left\| \bar{w}_{(n+j)N} - w^{T_n}(T_{n+j}) \right\|
\label{Eqn34}
\end{align}
Note $\mathrm{U}_n$ and $\mathrm{W}_n$
\begin{comment}
\begin{align}
\mathrm{U_{n}}
&= \left\| \sum_{j=0}^{q-1} \frac{1}{L} \sum_{i=1}^{L} \sum_{k=0}^{N-1} 
a^{(1)}_{(n+j)N+k} p^{(i)}_{(n+j)+k} 
M^{(i)}_{(n+j)N+k+1} \right\| \nonumber \\
&= \left\| \frac{1}{L} \sum_{i=1}^{L} 
\sum_{r=nN}^{(n+q)N-1} a^{(1)}_{r} p^{(i)}_{r} M^{(i)}_{r} \right\|
\label{Temp1}
\end{align}
and 
\begin{align}
\mathrm{W_{n}} 
&= \left\| \sum_{j=0}^{m-1} \frac{1}{L} \sum_{i=1}^{L} \sum_{k=0}^{N-1} 
a^{(1)}_{(n+j)N+k} \left( p^{(i)}_{(n+j)+k} - p^{(i)} \right) 
h^{(i)}\left(w^{(i)}_{(n+j)N+k}\right) \right\| \nonumber \\
&= \left\| \frac{1}{L} \sum_{i=1}^{L} \sum_{r=nN}^{(n+m)N-1} 
a^{(1)}_{r} \left( p^{(i)}_{r} - p^{(i)} \right) 
h^{(i)}\left(w^{(i)}_{r}\right) \right\|
\label{Temp2}
\end{align}
\end{comment}
converge to zero  as $n \to \infty$. 
Now, by discrete Grownwalls lemma 
\begin{align}
&\left\| \bar{w}_{(n+m)N} - w^{T_n}(T_{n+m}) \right\| \notag \\ 
&\quad \leq \left( \mathrm{P_{n}} + \mathrm{Q_{n}} + \mathrm{K_{n}}  + \mathrm{W_{n}} + \mathrm{U_{n}} \right) 
\exp\left( K_{Np} \sum_{j=0}^{m-1} a^{(1)}_{(n+j)N} \right).
\label{Eqn35}
\end{align}

Thus  $\parallel \bar{w}_{(n+m)N} - w^{T_n}(T_{n+m}) \parallel $ converges to zero as $n \to \infty$.
\section{Simulations}\label{Simulations}
This section presents two simulations that validate the theoretical analysis. The first simulation is on a synthetic linear regression problem, while the second involves training a classification model on the real-world image datasets. The first simulation does not involve a neural network and is provided here as a proof of concept to numerically validate the theoretical analysis. The code for both the simulations are available at~\url{https://zenodo.org/records/17082341}.

\subsection{Federated Linear Regression Setup} \label{FEDLINREG}
Consider a federated learning system with $L$ clients collaborating to learn a shared linear predictor. Each client $i$ generates samples
\[
\xi^{(i)}_k = (x^{(i)}_k, y^{(i)}_k) \in \mathbb{R}^d \times \mathbb{R}, \quad k=1,2,\ldots,n^{(i)},
\]
according to the linear model
\begin{align}
    y^{(i)}_k = x^{(i)\top}_k\, w^{(i)} + \epsilon^{(i)}_k,
    \label{REG1}
\end{align}
with $x^{(i)}_k \sim \mathcal{N}(0, (\sigma^{(i)}_x)^2 I_d)$, $\epsilon^{(i)}_k \sim \mathcal{N}(0, (\sigma^{(i)}_\epsilon)^2).$
Here $w^{(i)} \in \mathbb{R}^d$ denotes the true client-specific parameter vector. The design matrix $X^{(i)\top} \in \mathbb{R}^{n^{(i)} \times d}$ is formed by stacking rows $x^{(i)\top}_k$ and the response vector by stacking $y^{(i)}_k$, yielding
\begin{align}
    Y^{(i)} = X^{(i)\top} w^{(i)} + \epsilon^{(i)}.
    \label{REG3}
\end{align}
Let $f(w, \xi^{(i)}) \triangleq \left\| Y^{(i)} - X^{(i)\top} w \right\|^2$ and 
\begin{eqnarray}
    h^{(i)}(w) &\triangleq& - \nabla_w E_{\xi^{(i)}} \{ f(w, \xi^{(i)}) \}  = 2\, E_{\xi^{(i)}} \left[ X^{(i)}  (Y^{(i)} - X^{(i)\top}w) \right].
    \label{REG5}
  %  \\ \nonumber &=& 2\, E_{\xi^{(i)}} \left[ X^{(i)}  (Y^{(i)} - X^{(i)\top}w) \right].
\end{eqnarray}

%t of the expected loss 
%\begin{align}
  %  \nabla_w E_{\xi^{(i)}} \left\{ f(w, \xi^{(i)}) \right\} = -2\, E \left[ X^{(i)} X^{(i)\top} (w^{(i)} - w) \right].
   % \label{Eqn10}
%\end{align}
Clients perform stochastic gradient updates~\eqref{Eqn1} with
\begin{align}
S_m^{(i)} = \frac{1}{m} \sum^m_{k=1}  X^{(i)}  (Y^{(i)} - X^{(i)\top}w) ,  i = 1,2, \ldots L.
\label{REG12}
\end{align}
%is the mini-batch stochastic gradient estimate, unbiased with bounded variance per assumptions A4 and A5.
The weights $w^{(i)}_n$ converge to the zero of 
\begin{align}
    \frac{1}{L} \sum_{i=1}^L p^{(i)}h^{(i)}(w),
    \label{REG6}
\end{align}
which is   
\begin{align}
    w^* = \arg \min_{w \in \mathbb{R}^d} \sum_{i=1}^L p^{(i)} \, E \left\| Y^{(i)} - X^{(i)} w \right\|^2.
    \label{REG7}
\end{align}
Note
\begin{align}
     w^* =\left(\sum_{i=1}^L p^{(i)}\,\mathbb{E}[X^{(i)\top}X^{(i)}]\right)^{-1}\; \left(\sum_{i=1}^L p^{(i)}\,\mathbb{E}[X^{(i)\top}Y^{(i)}]\right).
    \label{REG8}
\end{align}

\subsubsection{Simulation Setup}\label{SimData}

The federated linear regression model \eqref{REG1} is simulated for $L=10$ clients. Each client’s features $x^{(i)}_k \in \mathbb{R}^3$ and parameters $w^{(i)} \in \mathbb{R}^3$ are sampled from $\mathcal{N}(0, 5^2I_3)$, while the additive noise $\epsilon^{(i)}_k$ is chosen to achieve an SNR of $10$ dB. Each client holds $n^{(i)} = 5000$ samples. Mini-batch size is fixed at $m=50$.  Clients run the iterations \eqref{Eqn3}  with $m=50, \ N=5$ and initial condition $w_0^{(i)} \sim \mathcal{N}(0,20^2 I_3)$. The mini-batch  gradients are computed using  \eqref{REG12}.    Each simulation runs for $5000$ federated rounds. Unless explicitly stated all  simulations
in this subsection  use the above configuration.

Convergence is assessed by comparing the aggregated iterates $\bar{w}_{nN}$ \eqref{Eqn13} to the closed-form global optimum $w^*$ in~\eqref{REG8}.  The aggregated gradient norms $\parallel\sum^{L}_{i=1} p^{(i)}h^{(i)}(\bar{w}_{nN}) \parallel$ and the parameter errors $\parallel  \bar{w}_{nN} - w^*\parallel$ are also tracked. 

\subsubsection{Consequences of Step-Size Choices on Convergence} \label{reg_consq_step_size}

 In order to empirically illustrate the influence of the step size ratios  $p^{(i)}$,  three  different cases are considered. 
 
\begin{figure}[!htb]
    \centering
    \includegraphics[width=0.8\textwidth]{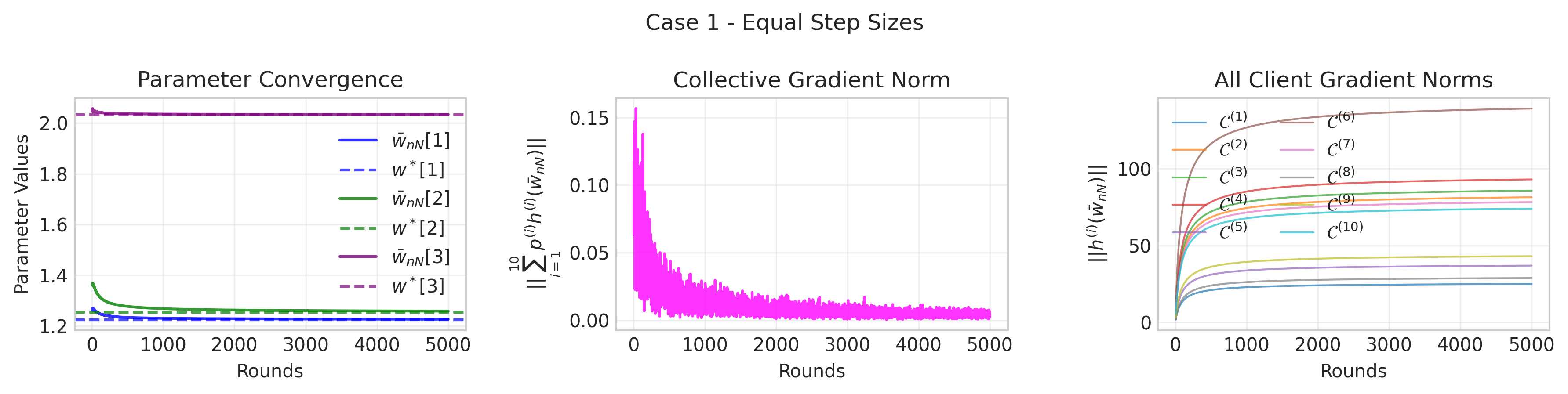}
    \caption{Equal Step Sizes. The parameters converge to the global minima. The aggregated gradient norm decays to zero and individual gradient norms are non-zero.}
    \label{fig3}
\end{figure}

\begin{figure}[!htb]
    \centering
    \includegraphics[width=0.8\textwidth]{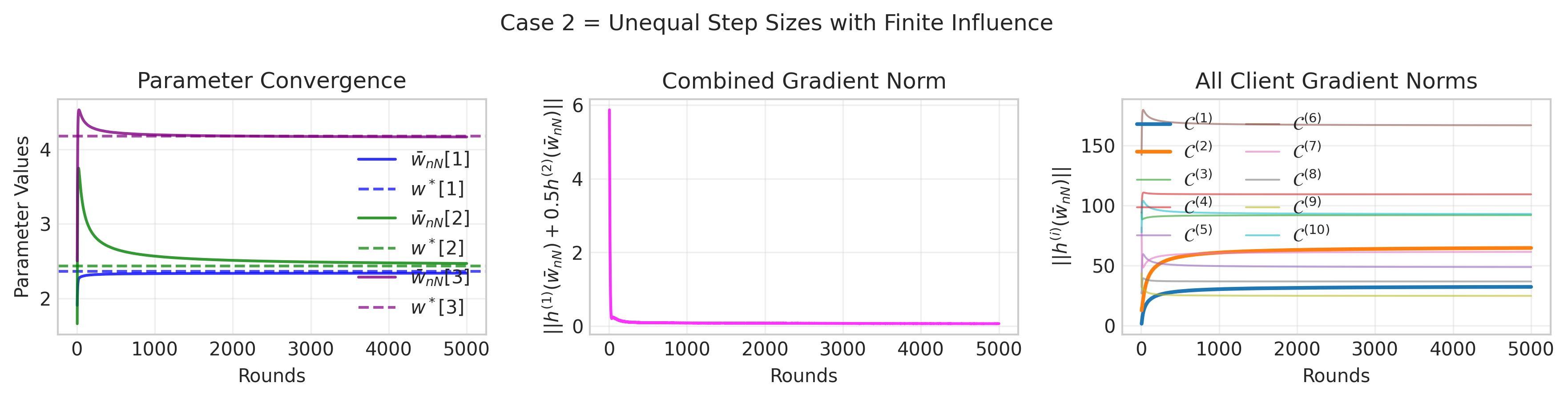}
    \caption{Unequal Step Sizes with Finite Influence. Only two clients influence the model. The combined gradient norms of client 1 and client 2 goes to zero while individual gradient norms are non-zero.}
    \label{fig9}
\end{figure}

\begin{figure}[!htb]
    \centering
    \includegraphics[width=0.8\textwidth]{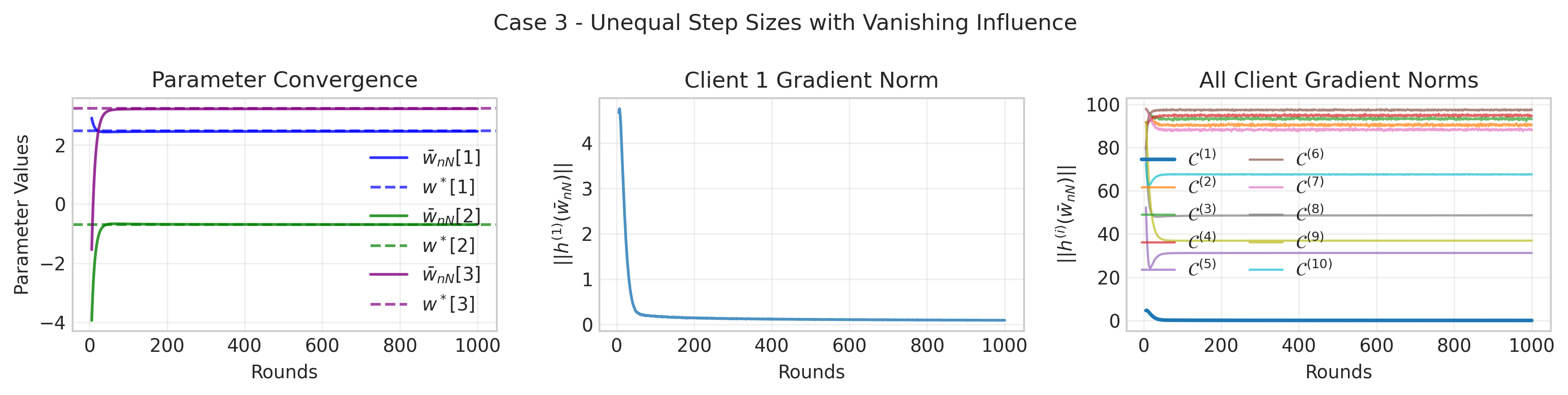}
    \caption{Unequal Step Sizes with Vanishing Influence. Only Client 1 has persistent influence. The global model converges to Client 1's parameters and only Client 1's gradient decay to zero.}
    \label{fig6}
\end{figure}

\paragraph{Case 1: Equal Step Sizes}
All clients use $a_n^{(i)} = 0.1/n^{0.76}$. This yields equal weights $p^{(i)}=1$.  The global optimal computed using \eqref{REG8}  turns out to be  $w^*=[1.22,1.25,2.03]$. From Figure~\ref{fig3} it can seen that $\bar{w}_{nN}$ converges to $w^*$,  while the aggregated gradient norm decays to zero.  Individual gradient norms $\parallel h^{(i)}(\bar{w}_{nN}) \parallel$ remain non-zero indicating that the  clients do not minimize their own local losses.
\paragraph{Case 2: Unequal Step Sizes with Finite Influence}
Clients 1 and 2 adopt slower decays ($a_n^{(1)} = 0.1/n^{0.76}, \ a_n^{(2)} = 0.1/(2n^{0.76})$), while others use $a_n^{(i)}=1/n$. Thus $p^{(1)}=1, p^{(2)}=0.5$, and others vanish. The global solution becomes a weighted combination of these two clients. 
Here, the optimal computed  using \eqref{REG8} turns out to be $w^*=[2.36,2.43,4.17]$. Parameter trajectories (Figure~\ref{fig9}) and  weighted  gradient norm $\parallel h^{(1)}(\bar{w}_{nN}) +0.5h^{(2)}(\bar{w}_{nN})\parallel$ confirm convergence.  Individual  clients gradients remain non-zero.
\paragraph{Case 3: Unequal Step Sizes with Vanishing Influence}
Only Client 1 maintains a slower decay $a_n^{(1)}=0.1/n^{0.76}$, others use $a_n^{(i)}=0.1/n, i \neq 1$. Thus $p^{(1)}=1$ and $p^{(i)} = 0, i \neq 1$. The global model converges entirely to Client 1’s optimum $w^*=[2.48,-0.69,3.23]$ and the Client 1’s gradient converges to zero, while others retain a non-zero value,  Figure.~\ref{fig6}.
\subsubsection{Variations in step size rates} \label{reg_tap_step_size}
\begin{figure}[!htb]
    \centering
    \includegraphics[width=0.6\textwidth]{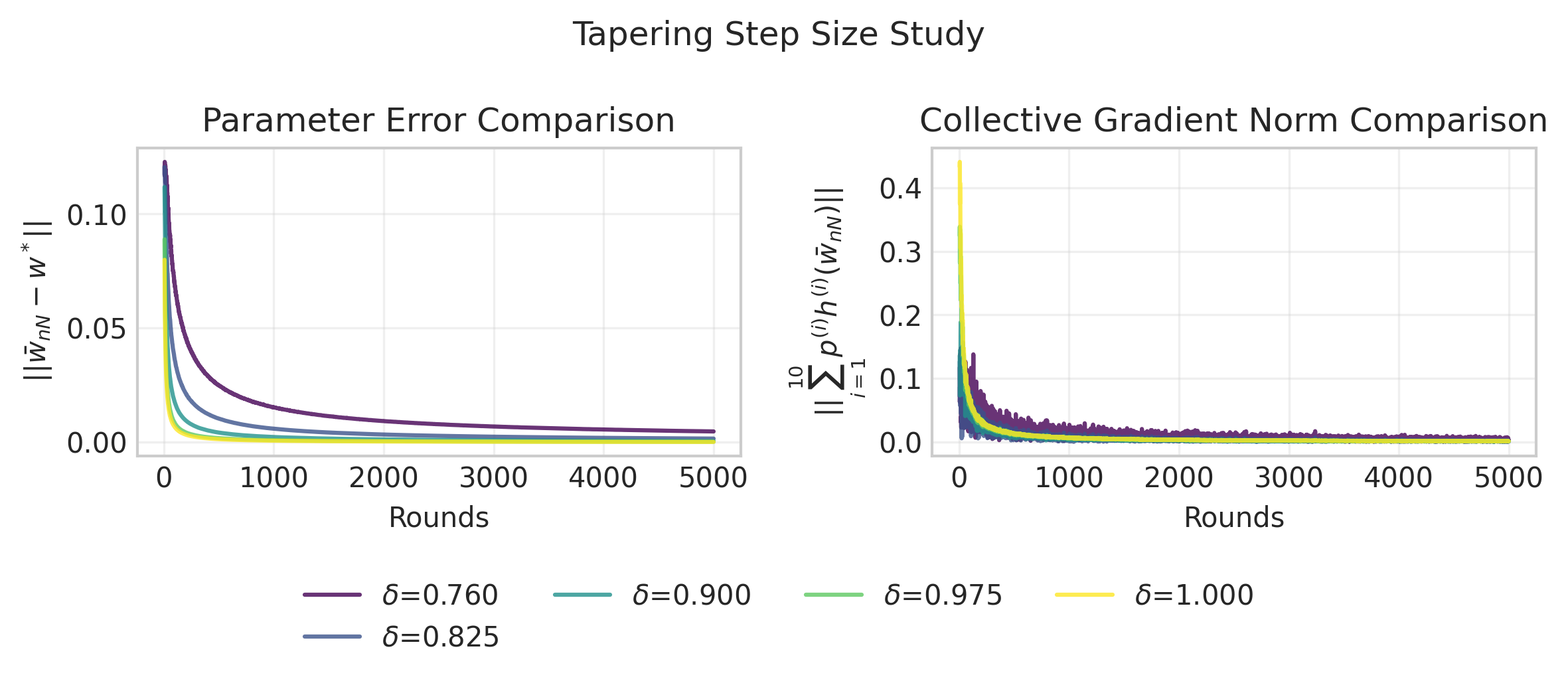}
    \caption{Effect of $\delta$ on parameter convergence and gradient norm decay. Smaller $\delta$ has faster decay but less stable convergence.}
    \label{fig14}
\end{figure}

The step size  is set to $a^{(i)}_n=0.1/n^\delta \ \forall i$. The effect of the variation in the  exponent $\delta$  is explored.   Here $\delta$ takes values in  $\{0.76,0.825,0.9,0.975,1.0\}$. 
Figure~\ref{fig14} shows that for smaller $\delta$ the aggregated gradient norm has a faster  rate of  convergence but results in oscillations, while larger $\delta$ leads to robust  convergence at the cost of a slower decay.  Due to oscillations the parameter error convergence is slower for smaller $\delta$ . 
\subsubsection{Client Heterogeneity and Varying model aggregation frequencies}\label{reg_data_het}
This study analysis three sources of heterogeneity for three different model aggregation frequencies $N \in \{ 2, 5, 10\}$:  

\textbf{1. Heterogeneity due to varying input feature distribution in the clients.} 
Here, heterogeneity is introduced by assigning each client  $i$  with a  input feature distribution.
$x^{(i)} \sim \mathcal{N}(0, (\sigma_x^{(i)})^2 I_3)$ with each   $\sigma^{(i)}_x$  randomly chosen from the set $\{5, 10, 15, 20, 25\}.$  The model parameter is fixed across all clients  to  \( w^{(i)} = (1.001, 0.998, 0.997)^\top \) and additive noise is generated to yield an SNR of 10 dB.  The step sizes are set to $ a_n^{(i)} = 0.1/n^{0.76}  \ \forall i$.

\begin{figure}[!htb]
    \centering
    \includegraphics[width=0.8\textwidth]{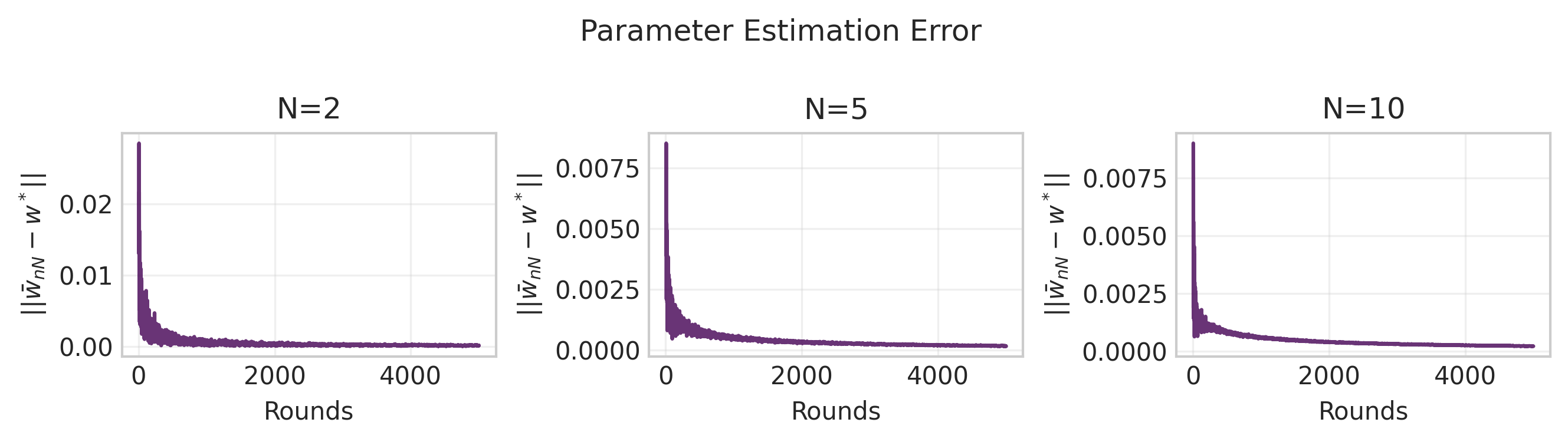}
    \caption{Parameter error under feature heterogeneity. Despite the heterogeneity, the parameters error decays to zero.}
    \label{figure18}
\end{figure}

\begin{figure}[!htb]
    \centering
    \includegraphics[width=0.8\textwidth]{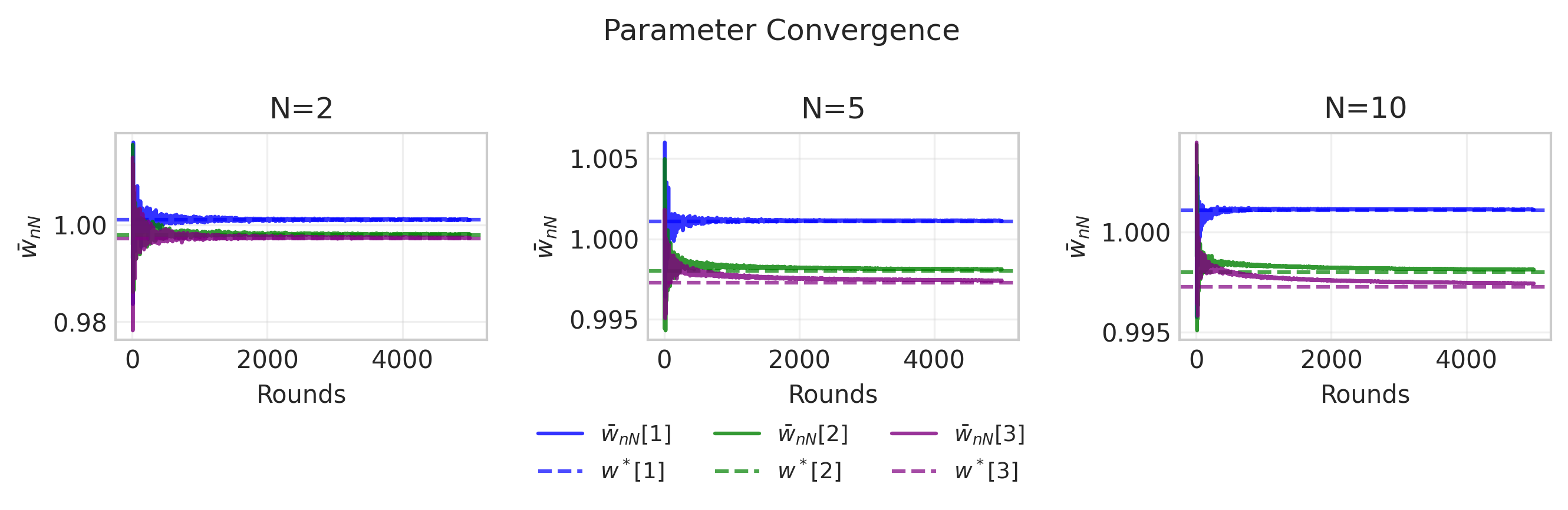}
    \caption{Parameter convergence under feature heterogeneity. The parameters converge to the desired values.}
    \label{figure19}
\end{figure}

\begin{figure}[!htb]
    \centering
    \includegraphics[width=0.8\textwidth]{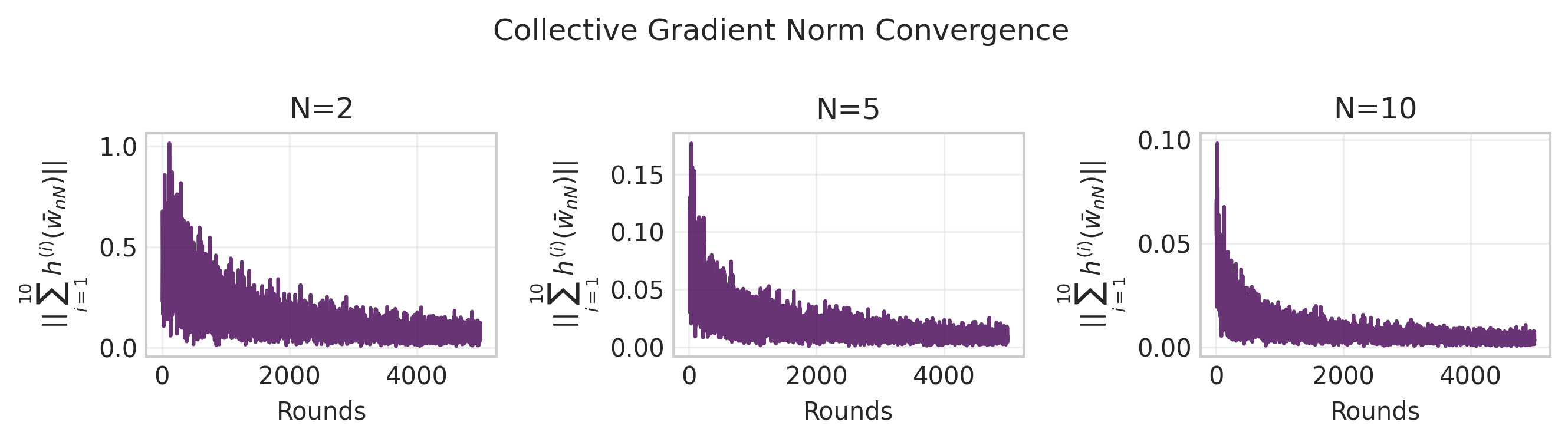}
    \caption{Aggregated Gradient norms under feature heterogeneity.}
    \label{figure20}
\end{figure}

In Figures~\ref{figure18} and \ref{figure19}, parameter estimation error and the parameter convergence trajectories are presented, while Figure~\ref{figure20} shows the corresponding aggregated gradient norm. With feature heterogeneity, the convergence is observed to become slower and more oscillatory. Despite the heterogeneity, the parameters converge to the desired values  (confirming the  probability  one convergence). 
 
\textbf{2. Heterogeneity due to larger variation in the client true parameters.} 
In all prior experiments  (except in the case of  feature heterogeneity), model parameters \( w^{(i)} \) were sampled independently for each client from \( \mathcal{N}(0, \sigma^2_wI_3 ) \) with $\sigma_w =5$.  Here, the effect of increasing  $\sigma_w$ is investigated.   The client parameter $w^{(i)}$ is sampled  for   $\sigma_w \in \{5, 10, 15, 20, 25\}$.  Here, the step sizes are set to \( a_n^{(i)} = 0.1/n^{0.76} \ \forall i\).

\begin{figure}[!htb]
    \centering
    \includegraphics[width=0.8\textwidth]{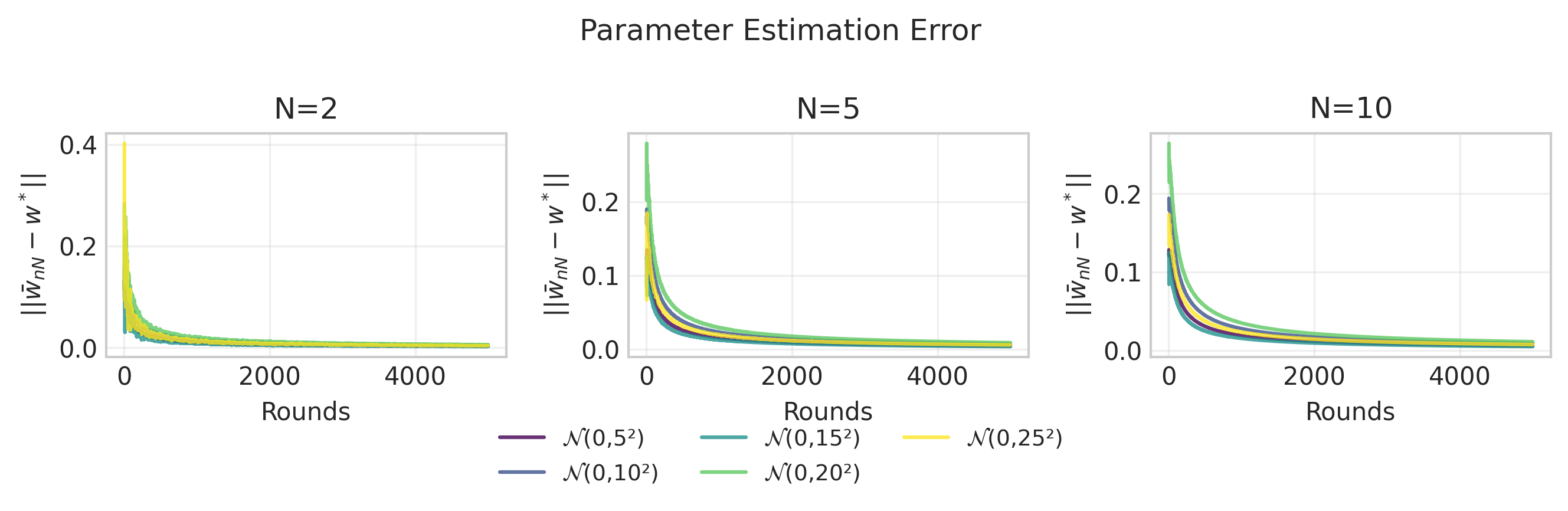}
    \caption{Parameter Estimation error under increasing $\sigma_w$.}
    \label{figure14p}
\end{figure}

\begin{figure}[!htb]
    \centering
    \includegraphics[width=0.8\textwidth]{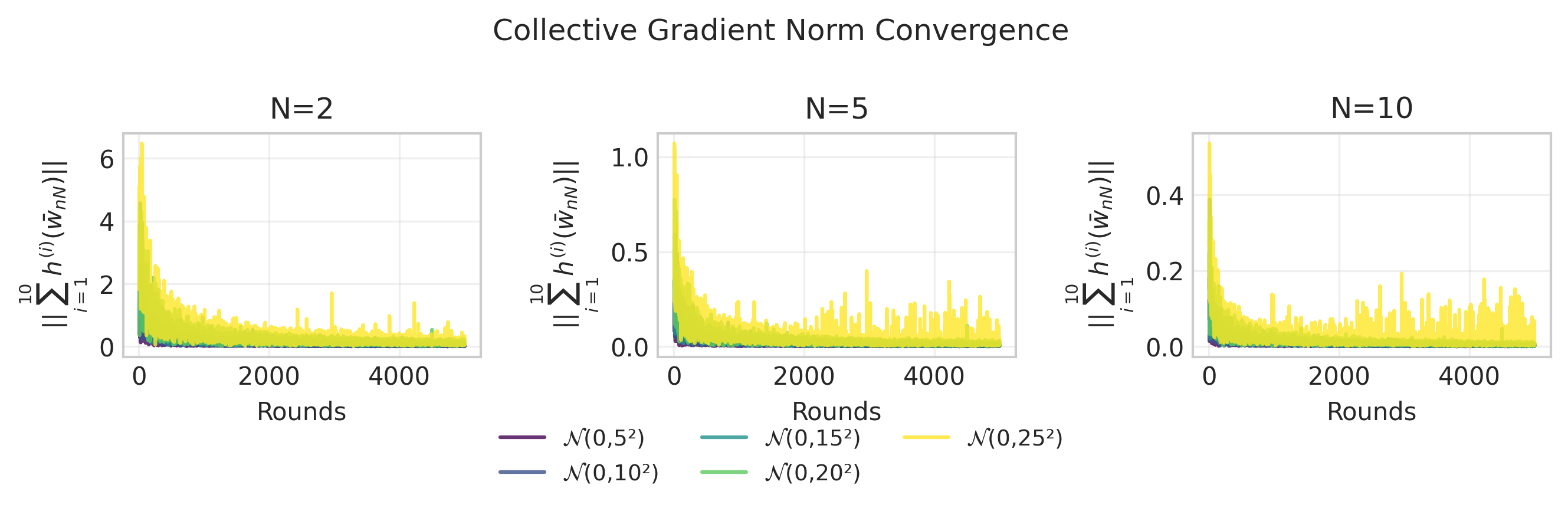}
    \caption{Aggregated Gradient norms under increasing $\sigma_w$.}
    \label{figure15p}
\end{figure}

In Figures~\ref{figure14p} and~\ref{figure15p}, parameter estimation error and the gradient norm trajectories, respectively, are plotted. As $\sigma_w$ increases, more persistent error and slower convergence are observed. This effect arises from the increasing divergence between local objectives, thereby reducing the effectiveness of global averaging in aligning the model updates.

\begin{figure}[!htb]
    \centering
    \includegraphics[width=0.8\textwidth]{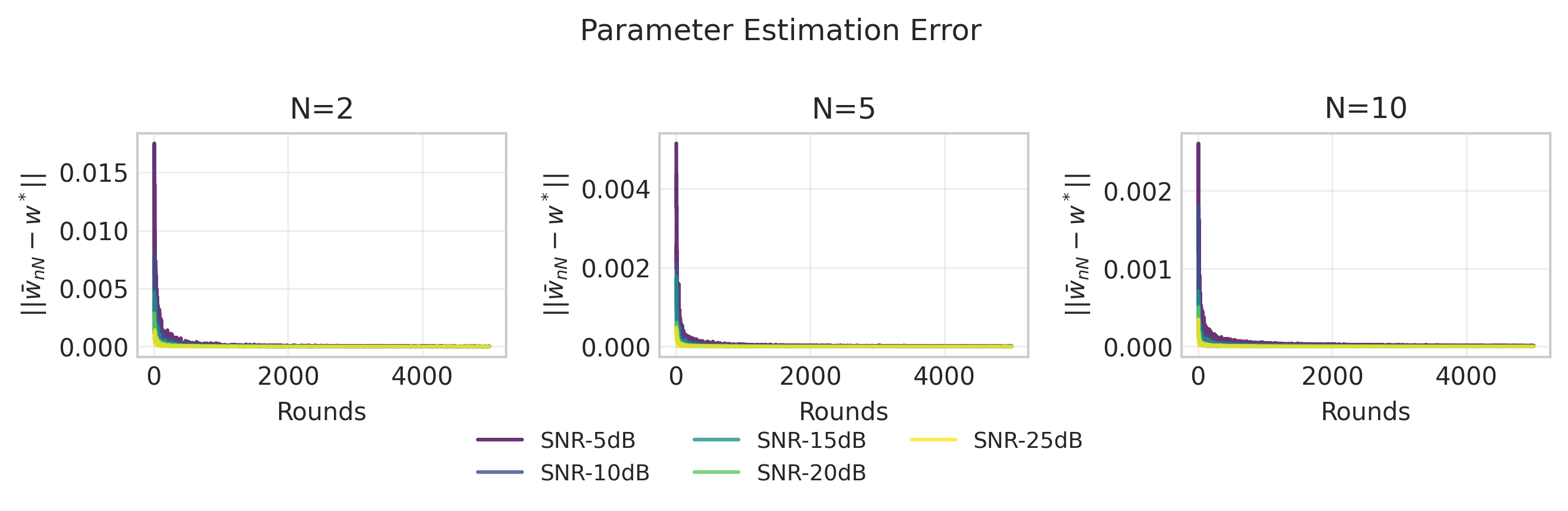}
    \caption{Parameter estimation error with varying levels of SNR.}
    \label{figure22}
\end{figure}

\begin{figure}[!htb]
    \centering
    \includegraphics[width=0.8\textwidth]{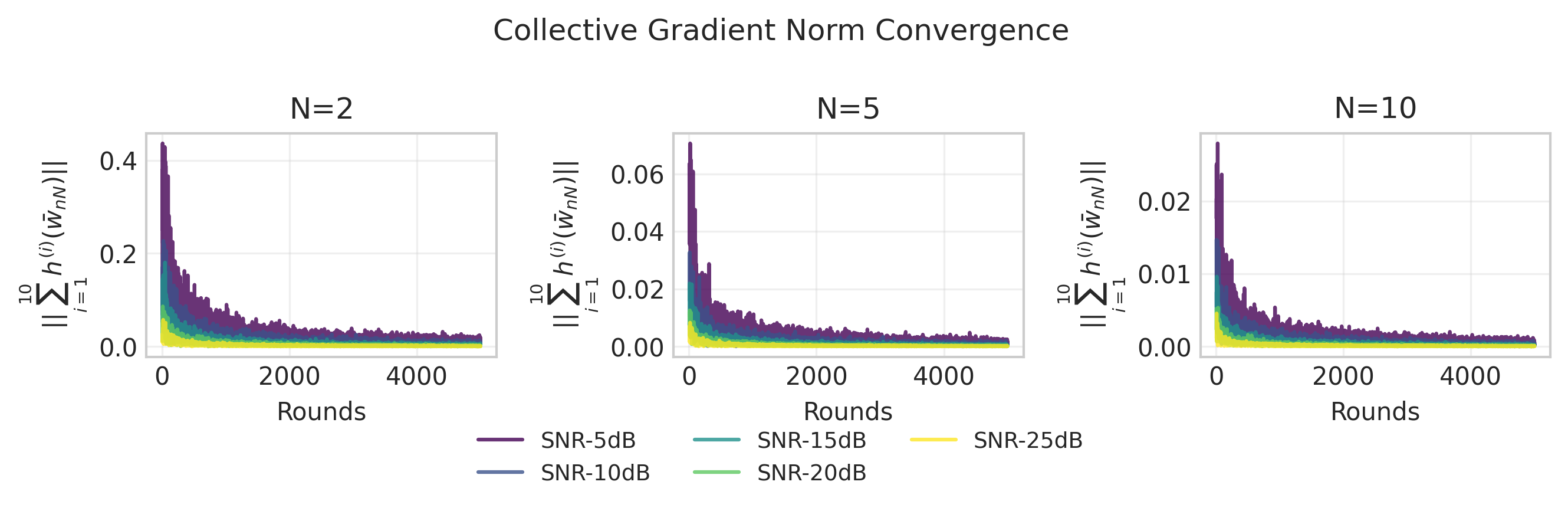}
    \caption{Aggregated Gradient norms convergence under varying levels of SNR.}
    \label{figure23}
\end{figure}

\textbf{3. Heterogeneity due to induced noise in the client data.}
In this setting, noise levels in the clients are varied by adjusting the signal-to-noise ratio (SNR).  The parameters are fixed to $ w^{(i)} = (1.001, 0.998, 0.997)^\top $ and the step sizes are set to \( a_n^{(i)} = 0.1/n^{0.76} \ \forall i\). The additive noise for the clients is varied across experiments to simulate decreasing SNR levels:
\[
\text{SNR} \in \{25\, \text{dB}, 20\, \text{dB}, 15\, \text{dB}, 10\, \text{dB}, 5\, \text{dB}\}.
\]
Figures~\ref{figure22} and~\ref{figure23} show that as the SNR is reduced, the parameter error shows oscillations, and the gradient norms decrease more slowly. This confirms that excessive noise degrades the convergence, even when the feature and the parameter distributions remain fixed. However, it is apparent from the plots that even for a very poor SNR value $(5\, \text{dB})$ the parameters still converge. This  implies the proposed framework  works for all finite noise variances. 

{\bf Across all heterogeneity types, larger $N$ reduces oscillations.}


\subsubsection{Comparison with Baseline Federated Methods}\label{reg_comparison}
\begin{figure}[htb!]
    \centering
    % Metrics plots stacked vertically
    \begin{subfigure}[t]{0.8\textwidth}
        \includegraphics[width=\linewidth]{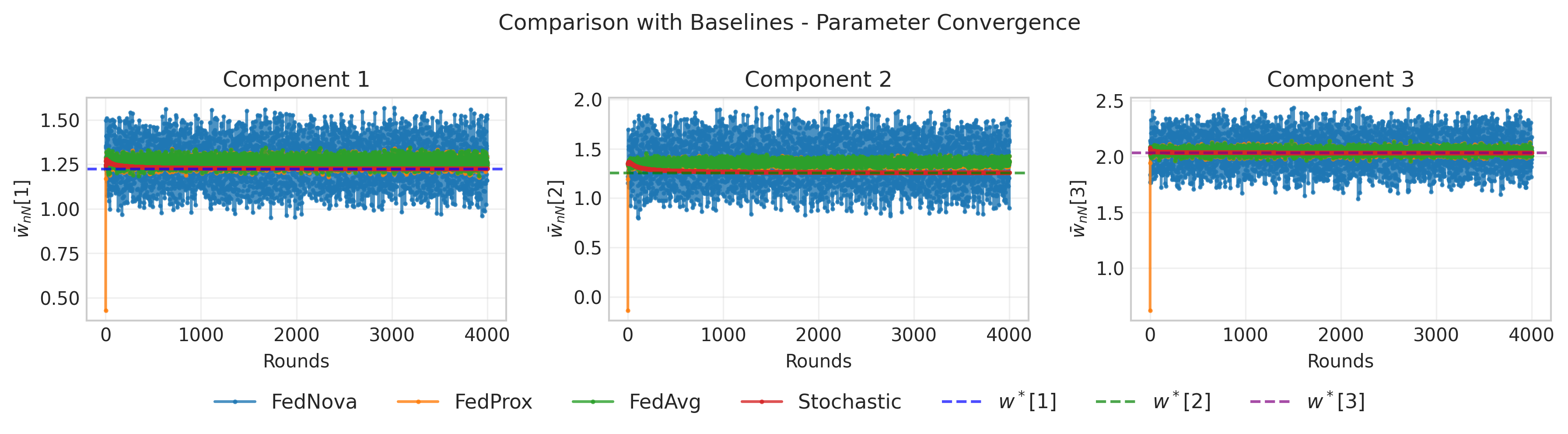}
        \caption{Constant step size: FedAvg, FedProx, FedNova do not converge.}
        \label{figure26a}
    \end{subfigure}
    
    \begin{subfigure}[t]{0.8\textwidth}
        \includegraphics[width=\linewidth]{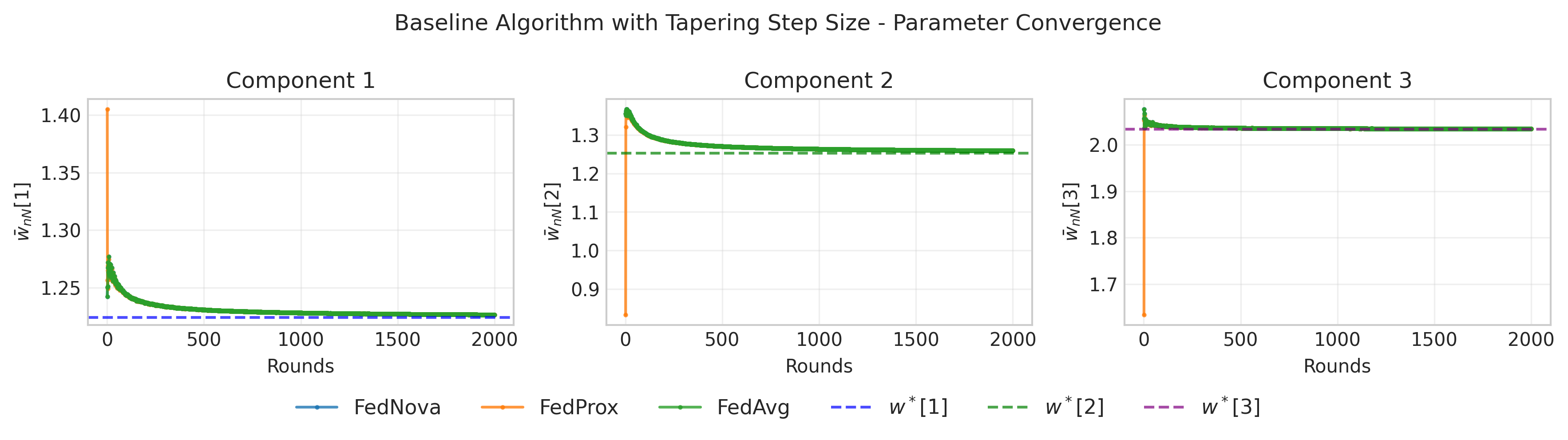}
        \caption{Tapering step size. FedAvg, FedProx, FedNova converge.}
        \label{figure26b}
    \end{subfigure}
    \caption{ Baseline algorithm comparison - Parameter convergence.}
\end{figure}

\begin{figure}[htb!]
    \centering
    % Row of gradient norm plots
    \begin{subfigure}[t]{0.4\textwidth}
        \includegraphics[width=\linewidth]{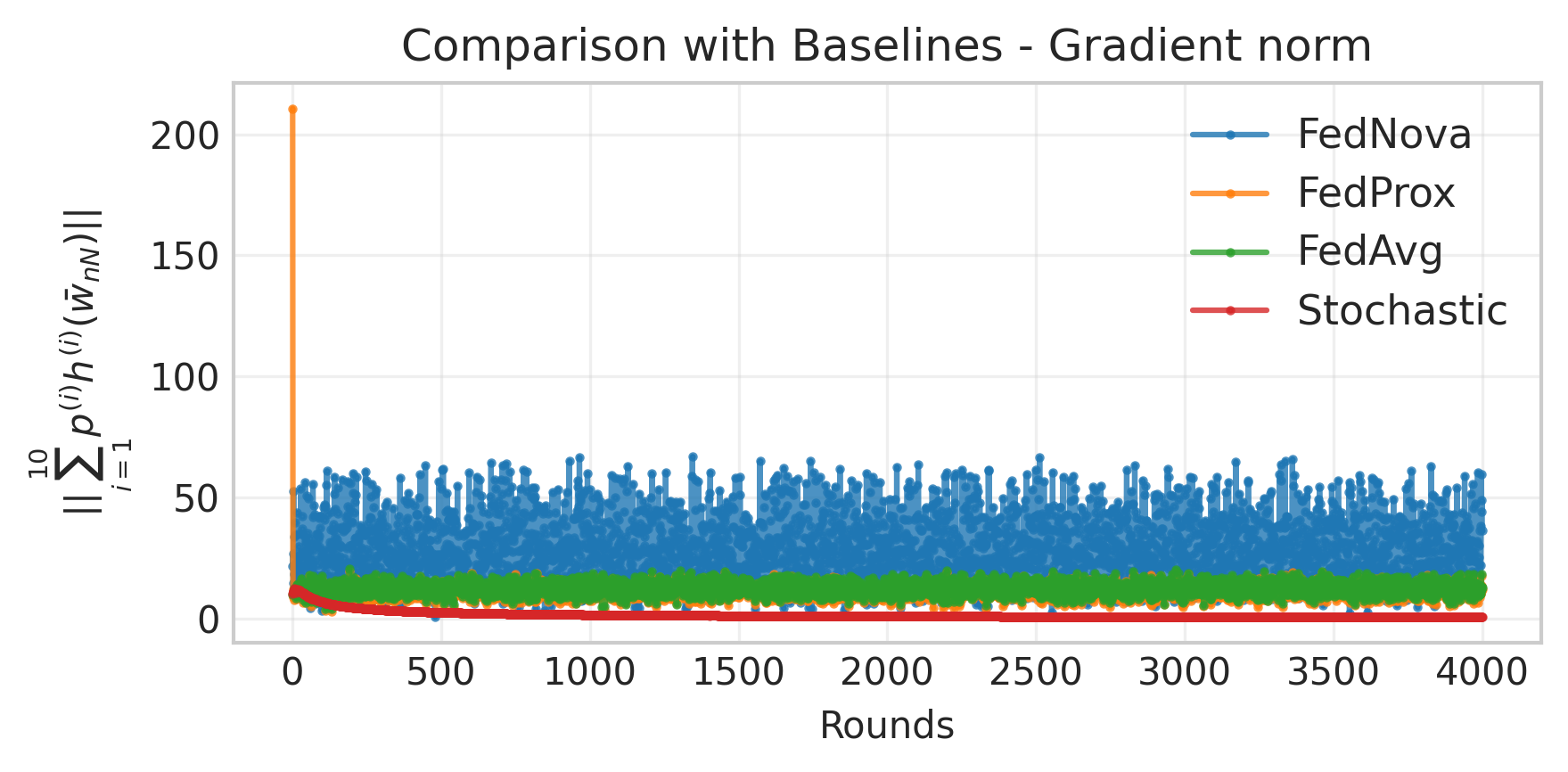}
        \caption{Constant step size: FedAvg, FedProx, FedNova gradient norms remain nonzero.}
        \label{figure27a}
    \end{subfigure}
    \begin{subfigure}[t]{0.4\textwidth}
        \includegraphics[width=\linewidth]{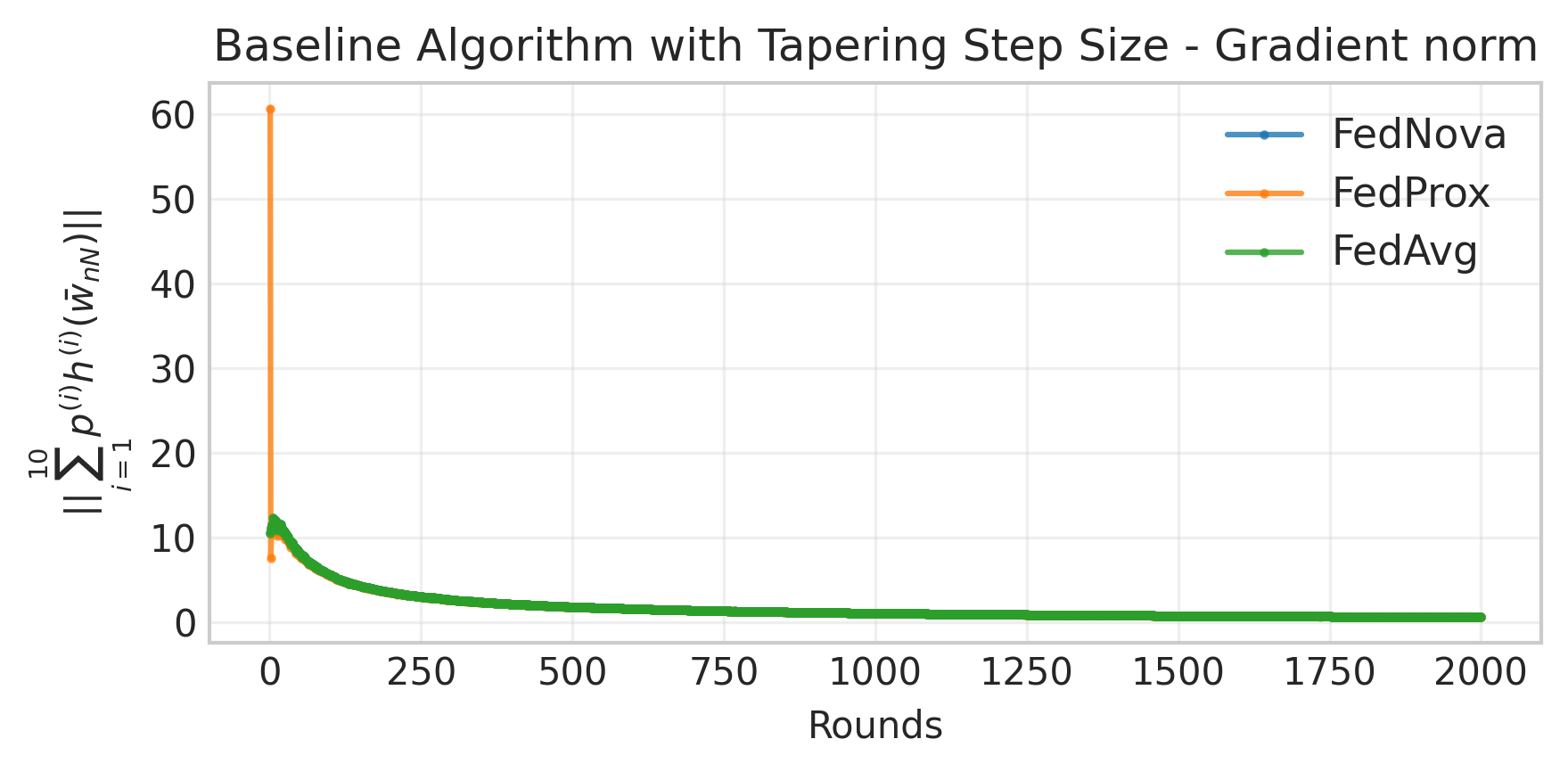}
        \caption{Tapering step size ensures FedAvg, FedProx, FedNova gradients decay to zero.}
        \label{figure27b}
    \end{subfigure}
    \caption{ Baseline algorithm comparison - Gradient norm.}
\end{figure}

The proposed method is compared with FedAvg, FedProx, and FedNova. Under constant step size  $a^{(i)}_n=0.1, \forall i$ convergence of the parameters and the aggregated gradient norm is plotted in Figures~\ref{figure26a}–\ref{figure27a}. It can be observed  that the baseline algorithms show oscillatory convergence. This is expected  as constant step-size algorithms converge only in expectation.   The proposed algorithm with  $a^{(i)}_n=0.1/n^{0.76}$, $\forall i$ shows robust convergence as the convergence is with probability 1.  However, if  the step sizes for the Baseline algorithms are set to   $a^{(i)}_n=0.1/n^{0.76}, \forall i$ then  their behavior is similar to the proposed algorithm as seen  from Figures.~\ref{figure26b}, \ref{figure27b}.  


\subsection{Federated Image Classification}\label{FEDIMGCLF}
Consider a federated $K$-class image classifier trained across $L$ clients. 
Client $i$ holds samples
\[
\xi^{(i)}_k=(x^{(i)}_k,y^{(i)}_k)\in\mathbb{R}^{H\times W\times C}\times\{1,\ldots,K\},\quad k=1,\ldots,n^{(i)},
\]
where $H\times W$ is the spatial size and $C\in\{1,3\}$ denotes the number of channels. 
Let $e(y)\in\{0,1\}^K$ denote the one‐hot encoding of label $y$. 
A client CNN with parameters $w\in\mathbb{R}^d$ (after flattening the weight tensors) produces logits $z^{(i)}(x;w)\in\mathbb{R}^K$. 
The corresponding softmax probabilities are
\begin{align}
s^{(i)}_r(x;w)=\frac{\exp\!\left(z^{(i)}_r(x;w)\right)}{\sum_{j=1}^K \exp\!\left(z^{(i)}_j(x;w)\right)}, 
\quad r=1,\ldots,K. 
\label{CE1-CE2}
\end{align}

The  cross-entropy loss  of  a sample  
\begin{align}
f\big(w, \xi^{(i)}_k\big) = - \sum_{r=1}^K e\big(y^{(i)}_k\big)_m \, \log s^{(i)}_r\big(x^{(i)}; w\big),
\label{CE3}
\end{align}
measures the discrepancy between the predicted probability vector $s(x; w)$ and the one-hot target vector $e(y)$.    The  loss function for each client $i$ is
\begin{align}
\mathcal{L}^{(i)}(w) \triangleq E_{\xi^{(i)}} \left\{ - \sum_{r=1}^K e(y)_k \, \log s^{(i)}_r(x^{(i)}; w) \right\}
\label{CE11}
\end{align}
and  the respective negative gradient   $h^{(i)}(w)= -\nabla\mathcal{L}^{(i)}(w)$.  And the global cost function to be minimized is 
\begin{align}
\mathcal{L}(w)=\sum_{i=1}^L p^{(i)}\mathcal{L}^{(i)}(w).
\label{CEagg}
\end{align}

Unlike regression here the number of parameters are very large and  the derivatives    are non-linear in the  parameter $w$.   Therefore  deriving or finiding an expression  for the zeros of 
\begin{eqnarray}
    \sum^L_{i=1} p^{(i)} h^{(i)}(w) &=& 0
    \label{CE13}
\end{eqnarray}
is not possible. 

\subsubsection{Simulation Setup} \label{CLF_SIM_CONF}
The federated image classification is evaluated with $L=10$ clients on three benchmark datasets   OrganSMNIST ($H{=}28$, $W{=}28$, $C{=}1$, $K{=}11$),  BloodMNIST ($H{=}28$, $W{=}28$, $C{=}3$, $K{=}8$) and CIFAR-10 ($H{=}32$, $W{=}32$, $C{=}3$, $K{=}10$). OrganSMNIST  is a medical imaging dataset consisting  of the scans of human organs in the sagittal-view \cite{yang2021medmnist}. BloodMNIST is a medical imaging dataset of blood cell microscopy images~\cite{yang2021medmnist} and CIFAR-10 is a collection of natural images~\cite{krizhevsky2009learning}.

To simulate a heterogeneous federated learning setting, we adopt a near-pathological non-identical partitioning scheme: for each client majority ($65$--$80\%$, depending on dataset) of the local data is sampled from a single dominant class, while the remainder is drawn uniformly from the other classes. All datasets are normalized with their dataset-specific mean and standard deviation. For BloodMNIST and OrganSMNIST additional augmentations such as random horizontal flips and small rotations are included during training to improve robustness.
%A LeNet-5 (ReLU; $\sim$44k params) architecture is used for MNIST while 

A ResNet-9 ( with GroupNorm, ReLU, $\sim$6.6M parameters) is used as the model architecture~\cite{He2015DeepRL}. A mini-batch SGD with batch size of $m=32$ is used as the local optimizer. Aggregation happens every $N=3$ local epochs. A total of $100$ federated rounds are performed. The Average client train loss/accuracy, the centralized or global test loss/accuracy and  the parameter error  $\Delta \bar{w}_{nN}=\|\bar{w}_{(n+1)N}-\bar{w}_{nN}\|$  are tracked. Note as the number of parameters are very large (6.6 Million)   determining $w^*$ by solving \eqref{CE13}  and also plotting the individual parameters and the gradients are not possible. However the convergence of $\Delta \bar{w}_{nN}$  along with convergences of training metrics and test metrics imply convergence to a local minima. Unless explicitly mentioned the following   numerical experiments use the above mentioned simulation configuration. 

Note here for brevity, the study on the effects of varying $\delta$,  varying $N$ and heterogeneity levels are presented in the  Appendix E of the Supplementary material.

\subsubsection{Consequences of Step-Size Choices on Convergence}
\begin{figure}[htb!]
    \centering
    \includegraphics[width=0.75\linewidth]{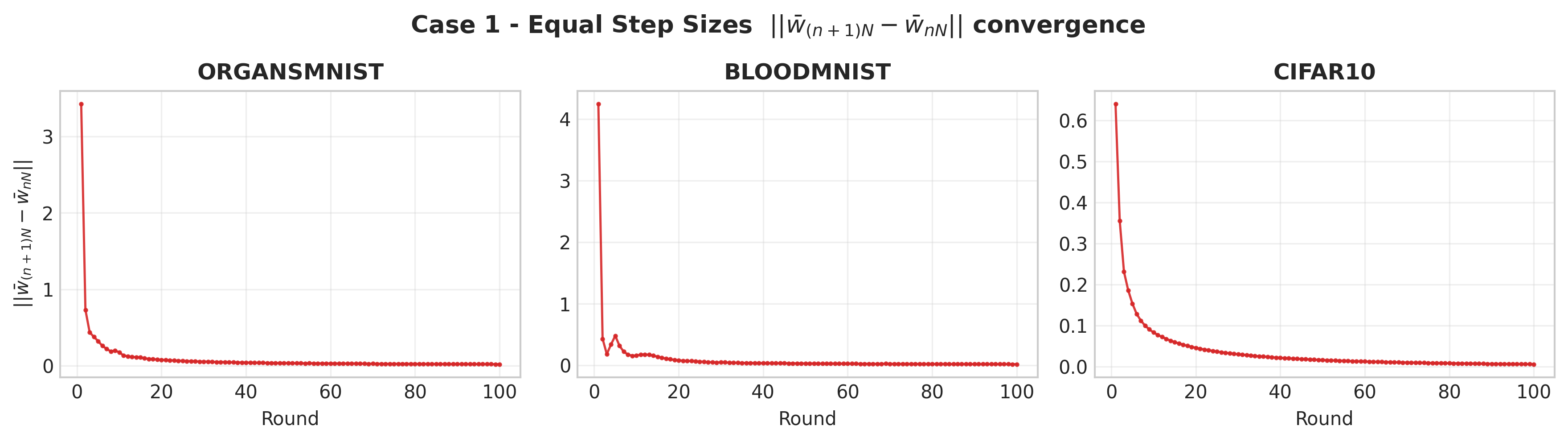}
    \caption{$\|\bar{w}_{(n+1)N}-\bar{w}_{nN}\|$ convergence across datasets under equal step sizes.}
    \label{fig:gradnorm_all}
\end{figure}

\begin{figure}[htb!]
    \centering
    \begin{subfigure}[t]{0.95\textwidth}
        \includegraphics[width=\linewidth]{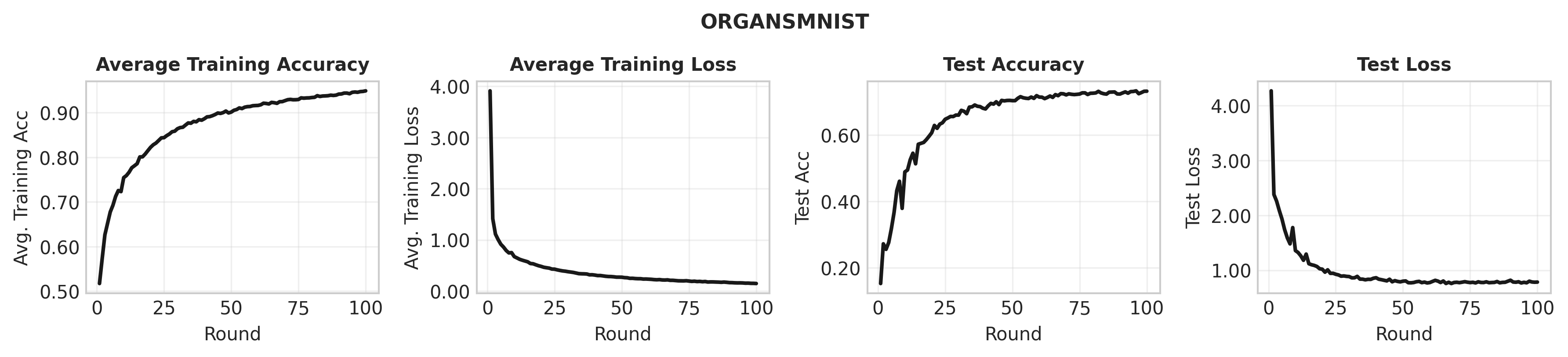}
        \caption{OrganSMNIST}
        \label{figure107b}
    \end{subfigure}
    \begin{subfigure}[t]{0.95\textwidth}
        \includegraphics[width=\linewidth]{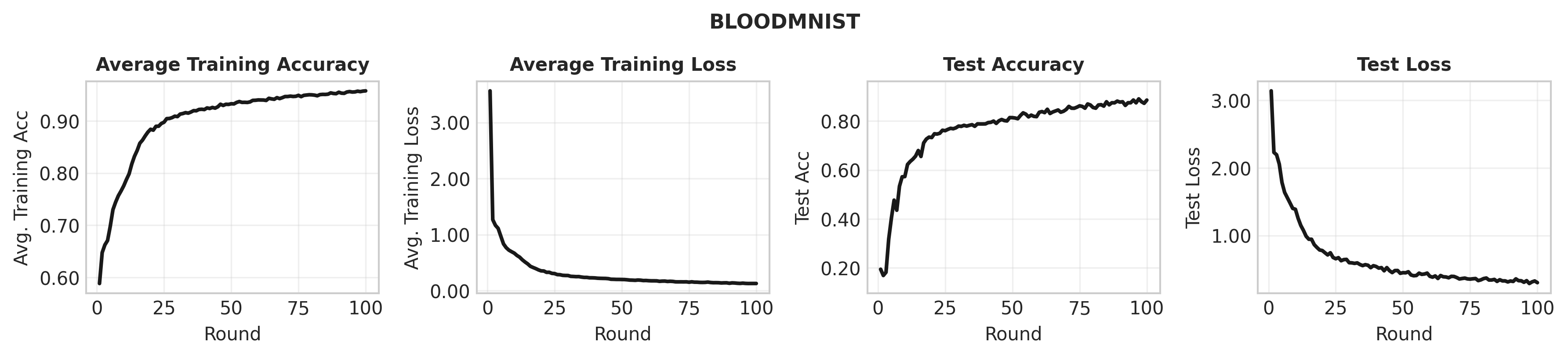}
        \caption{BloodMNIST}
        \label{figure114b}
    \end{subfigure}
    \begin{subfigure}[t]{0.95\textwidth}
        \includegraphics[width=\linewidth]{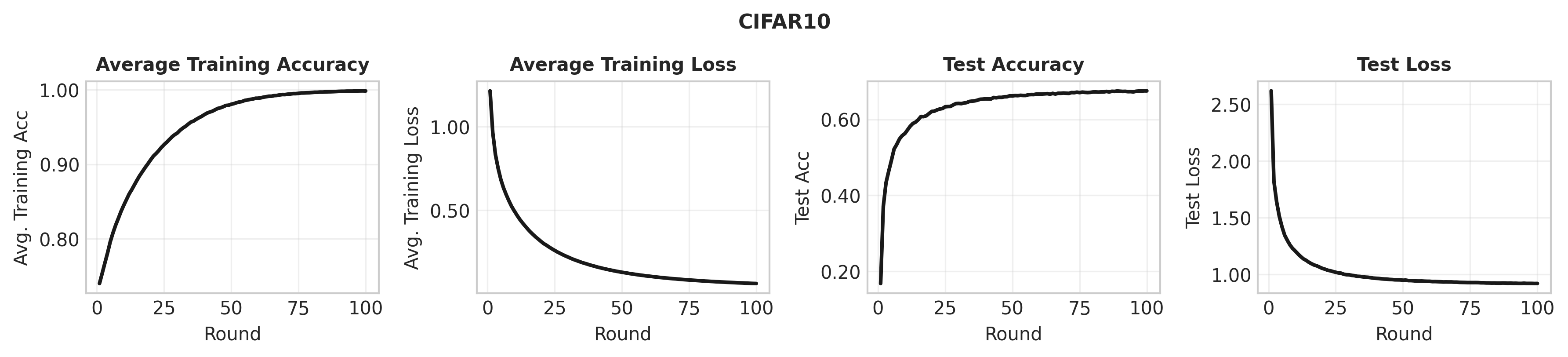}
        \caption{CIFAR-10}
        \label{figure121b}
    \end{subfigure}
    \caption{Average training and centralized test metrics under equal step sizes across datasets.}
    \label{fig:metrics_all}
\end{figure}

\textbf{Equal step sizes:} Here, the step sizes are set to $a_n^{(i)}=0.1/n^{0.76}$ for all $i$. This makes   $p^{(i)}{=}1$. In Figure~  \ref{fig:gradnorm_all} $\Delta \bar{w}_{nN}$  is plotted for all the three data sets.  It can be seen that  $\Delta \bar{w}_{nN}$ converges to zero.  Average training  and centralized test metrics for OrganSMNIST, BloodMNIST and CIFAR-10 are plotted  in  Figures.~\ref{figure107b}, \ref{figure114b} and\ref{figure121b} respectively. It can be observed that training  loss converges to zero while the  test loss decreases consistently for all the three data sets. Similarly training accuracy converges to one while test accuracy  improves consistently to competitive values, implying convergence to a local minima.  

\textbf{Unequal step sizes with Finite Influences:}   
Here, the data partitioning scheme differs slightly from that described in Subsection~\ref{CLF_SIM_CONF}.~Images from a Class~$r \in K$ are assigned exclusively to Client~1, while the remaining $K-1$ classes are distributed uniformly among the remaining clients. Accordingly, Client~1 is referred to as the \textit{rare-class client} and Class~$r$ as the \textit{rare class}. Note that the rare class considered is liver ($r=6$) in OrganSMNIST, neutrophil ($r=6$) in BloodMNIST and airplane ($r=0$) in CIFAR-10, following the datasets’ original label mappings.

\begin{figure}[htb!]
    \centering
    \begin{subfigure}[t]{0.45\textwidth}
        \includegraphics[width=\linewidth]{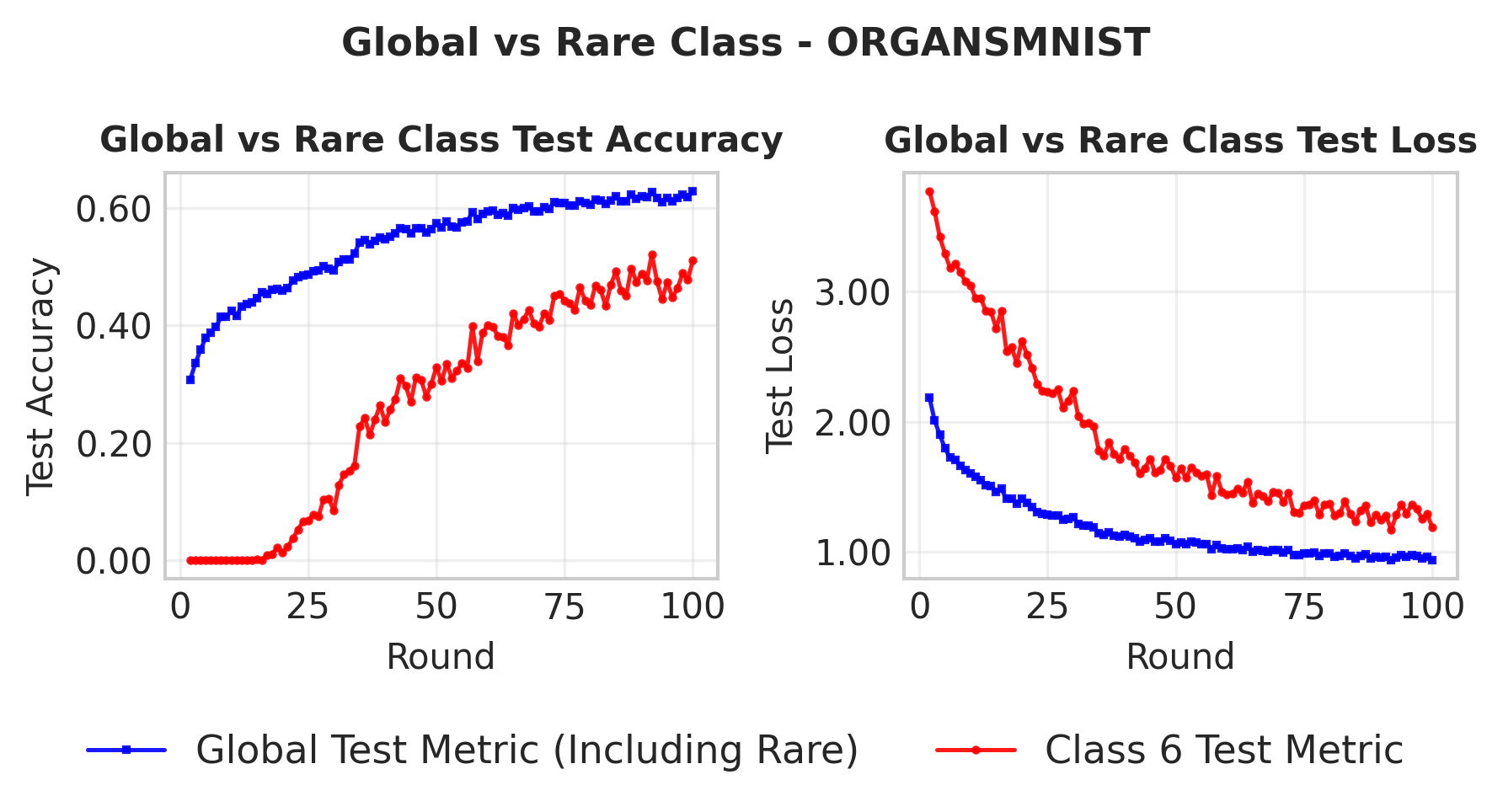}
        \caption{OrganSMNIST:  $p^{(i)}=1  \ \forall i$.}
        \label{figure109a}
    \end{subfigure}
    \begin{subfigure}[t]{0.45\textwidth}
        \includegraphics[width=\linewidth]{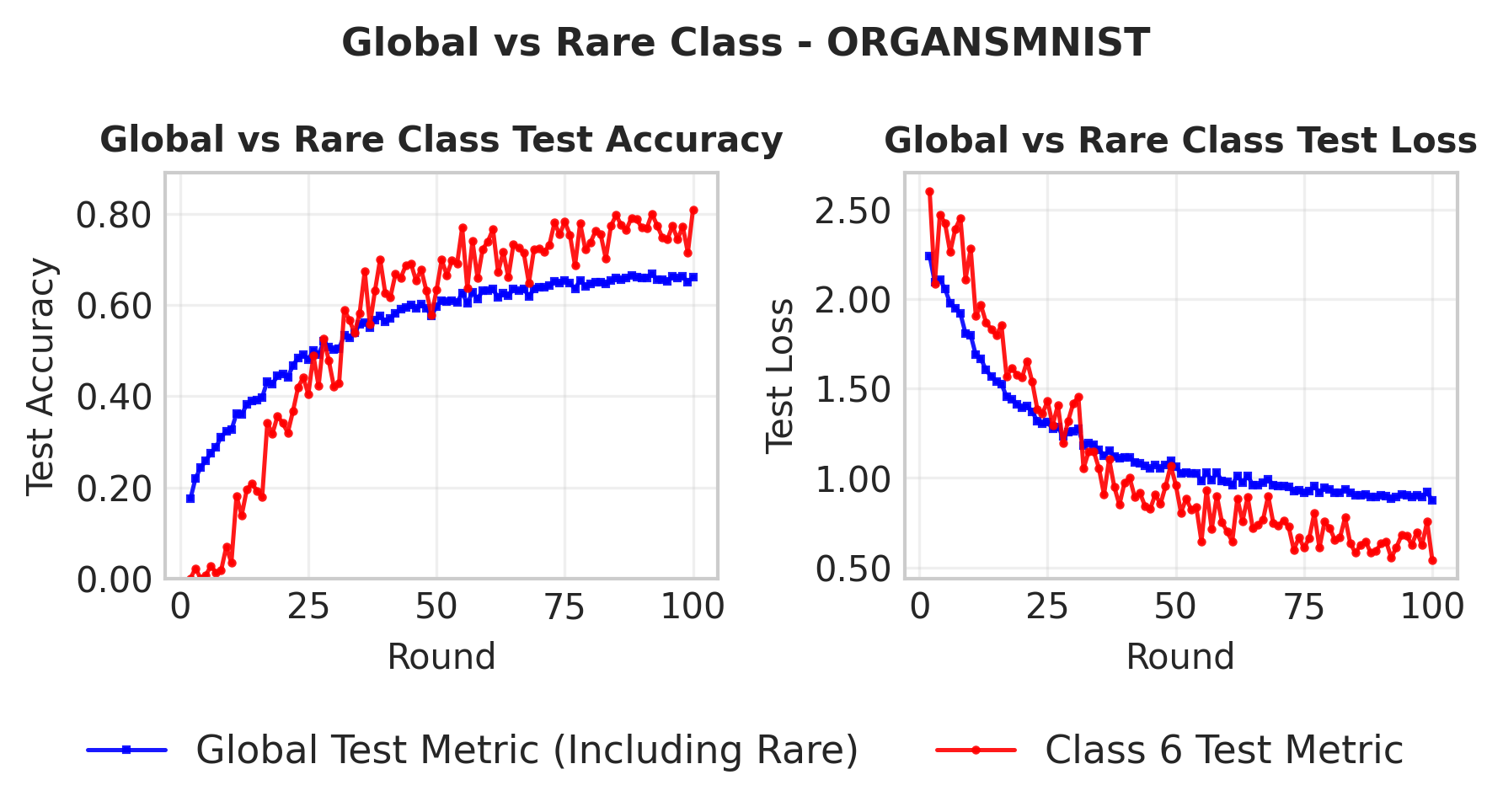}
        \caption{OrganSMNIST: $p^{(1)})= 1 \ p^{(i)}=0.1, i \neq 1$.}
        \label{figure109b}
    \end{subfigure}
    \begin{subfigure}[t]{0.45\textwidth}
        \includegraphics[width=\linewidth]{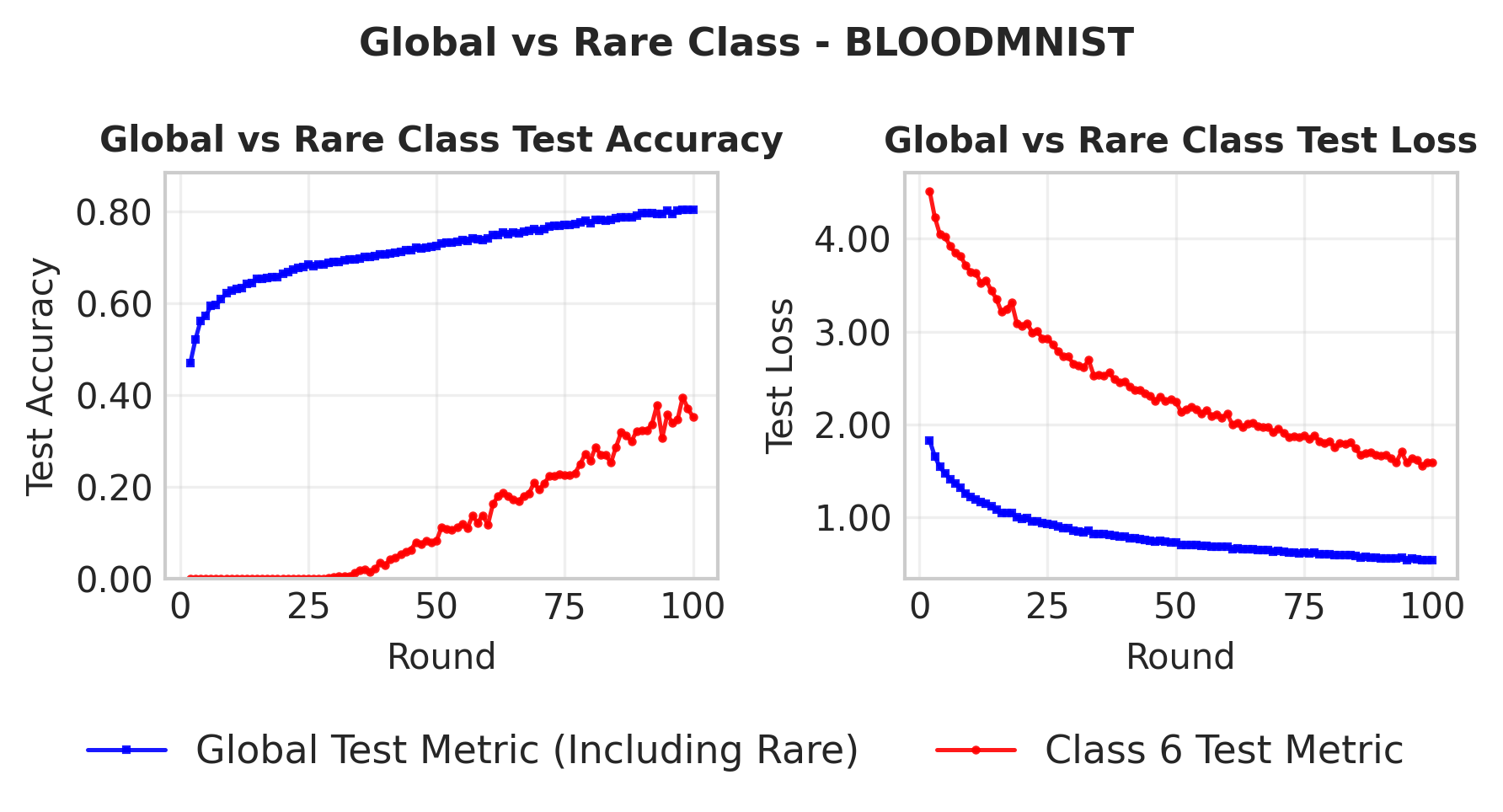}
        \caption{BloodMNIST: $p^{(i)} = 1 \ \forall i$.}
        \label{figure116a}
    \end{subfigure}
    \begin{subfigure}[t]{0.45\textwidth}
        \includegraphics[width=\linewidth]{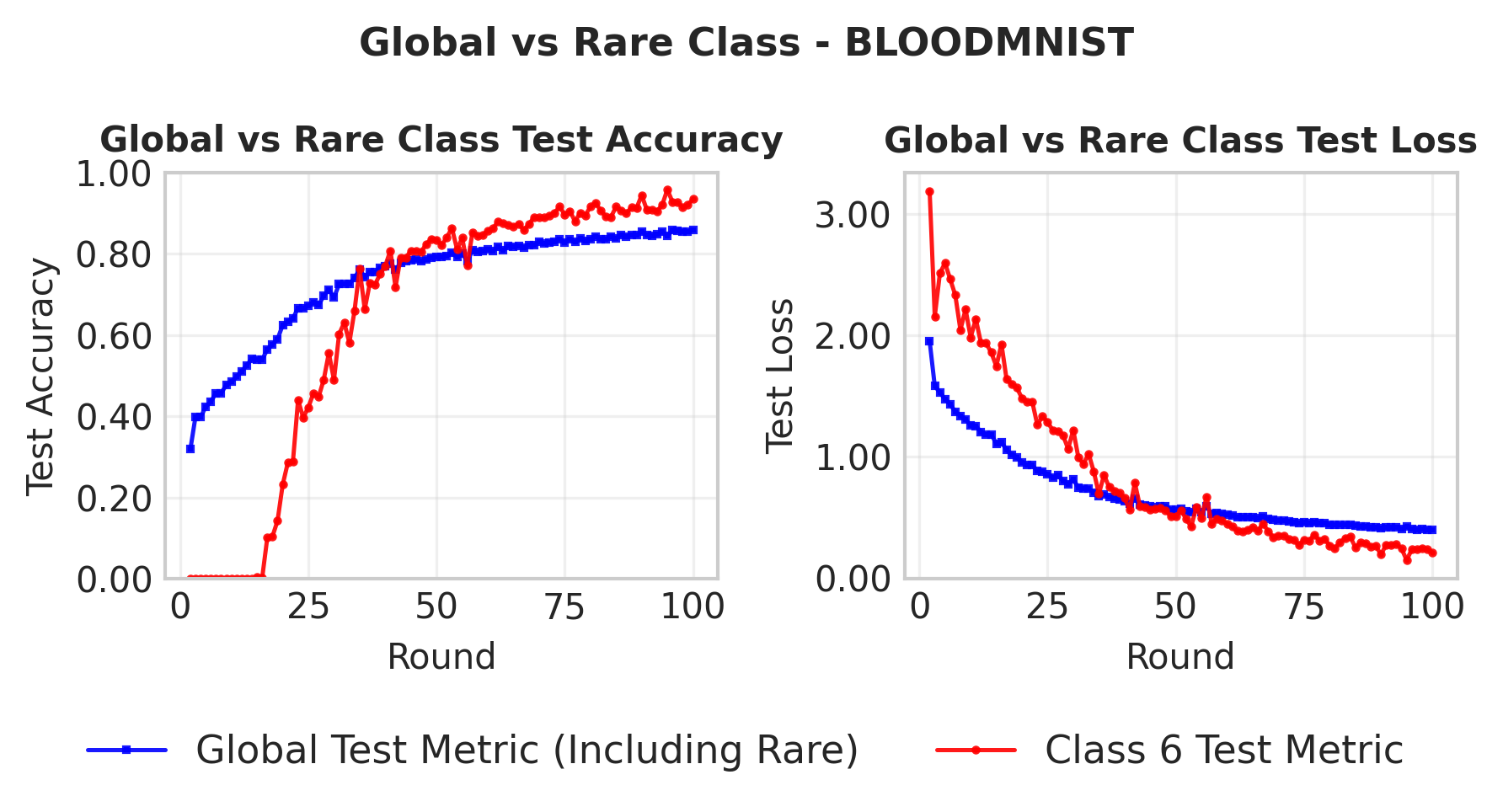}
        \caption{BloodMNIST: $p^{(1)})= 1 \ p^{(i)}=0.1, i \neq 1$.}
        \label{figure116b}
    \end{subfigure}
    \begin{subfigure}[t]{0.45\textwidth}
        \includegraphics[width=\linewidth]{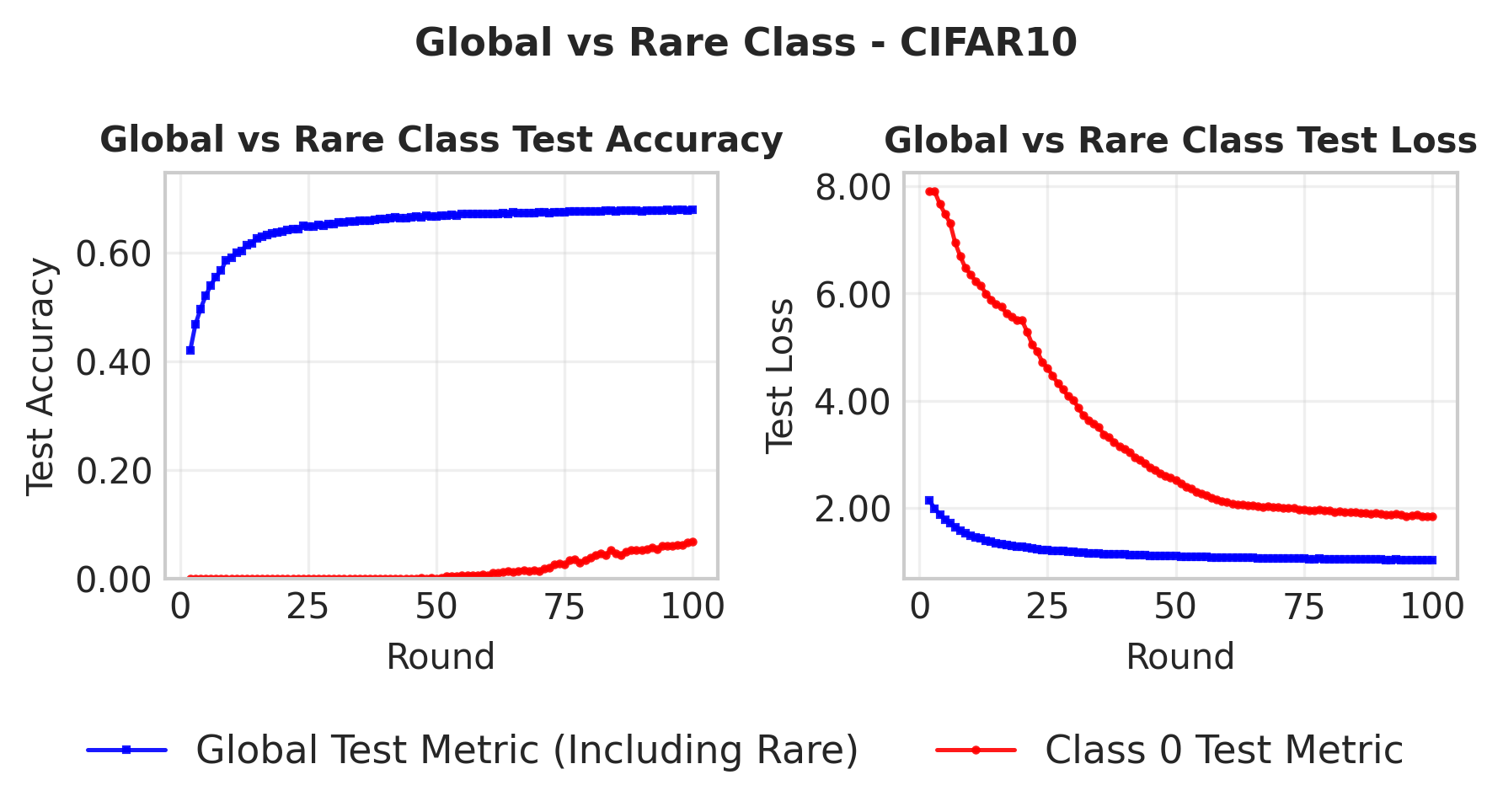}
        \caption{CIFAR-10: $p^{(i)} = 1 \ \forall i$.}
        \label{figure123a}
    \end{subfigure}
    \begin{subfigure}[t]{0.45\textwidth}
        \includegraphics[width=\linewidth]{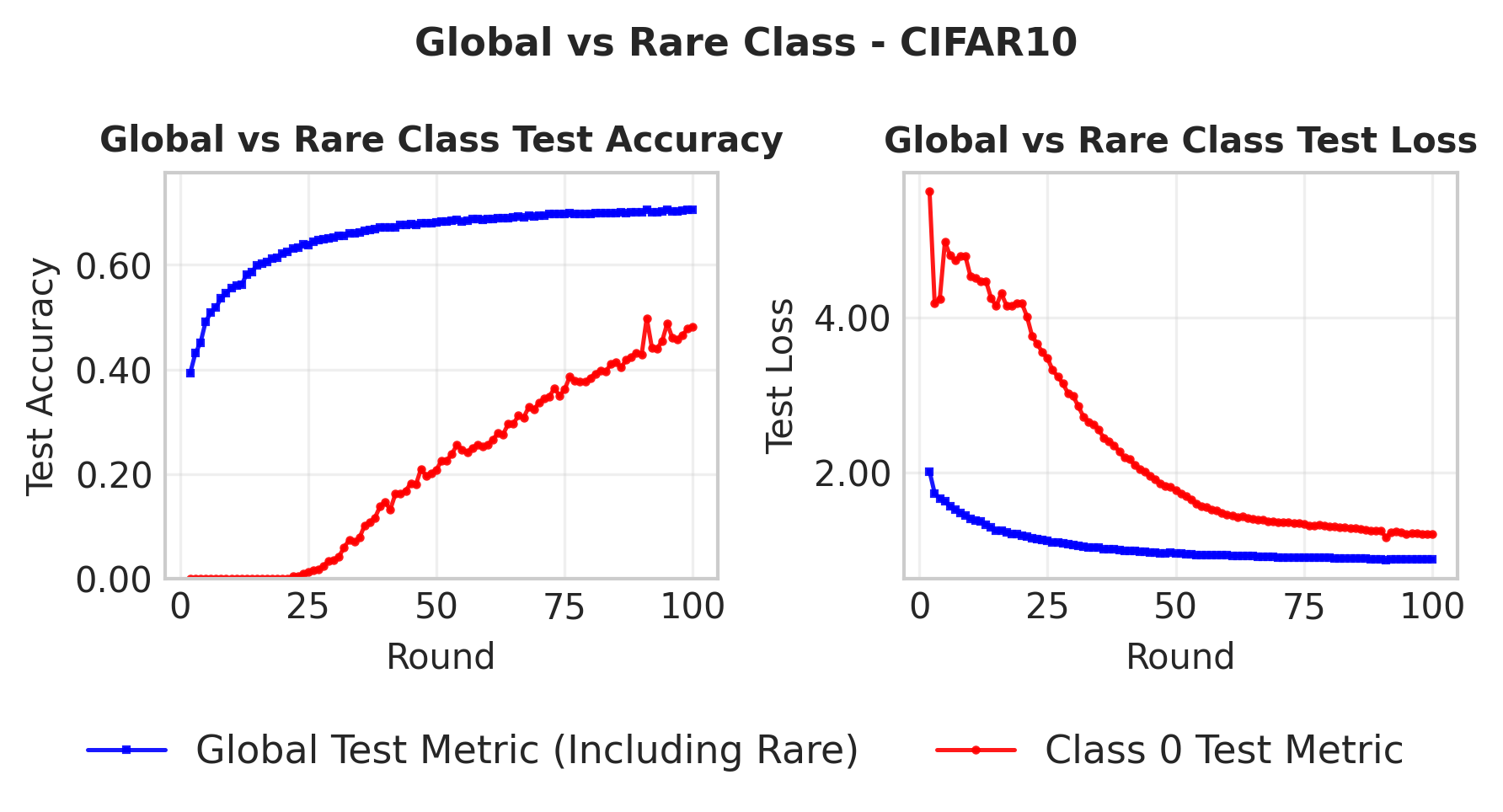}
        \caption{CIFAR-10: $p^{(1)})= 1 \ p^{(i)}=0.1, i \neq 1$.}
        \label{figure123b}
    \end{subfigure}

    \caption{Comparison of test performance when the rare client uses the same step size as others {\it i.e.,} $p^{(1)}=1$ (left) versus a higher step size  $p^{(1)}=1$ and $p^{(i)}=0.1$ for $i \neq 1$ (right).}
    \label{fig:case2_all}
\end{figure}

For Client~1  $a_n^{(1)} = 0.1/n^{0.76}$ and for all other clients
$a_n^{(i)} = 0.01/n^{0.76},\ i \neq 1$. Hence $p^{(1)} = 1$ and $p^{(i)} = 0.1$ for $i \neq 1$.  
To highlight the benefit of this design, Figures~\ref{figure109a}\ref{figure116a} and \ref{figure123a} shows the test loss and accuracy for all the three datasets when all clients use the same step size {\it i.e.,} $p^{(i)} = 1$ for all $i$. 
In contrast, Figure~\ref{figure109b},\ref{figure116b} and \ref{figure123b} illustrates the case with heterogeneous step sizes mentioned above, where $p^{(1)} = 1$ and $p^{(i)} = 0.1$ for $i \neq 1$. 
In the latter configuration rare-class test performance is consistently improved  while preserving competitive global accuracy.  

\begin{figure}[htb!]
    \centering
    % First row: two subfigures
    \begin{subfigure}[t]{0.49\textwidth}
        \centering
        \includegraphics[width=\linewidth]{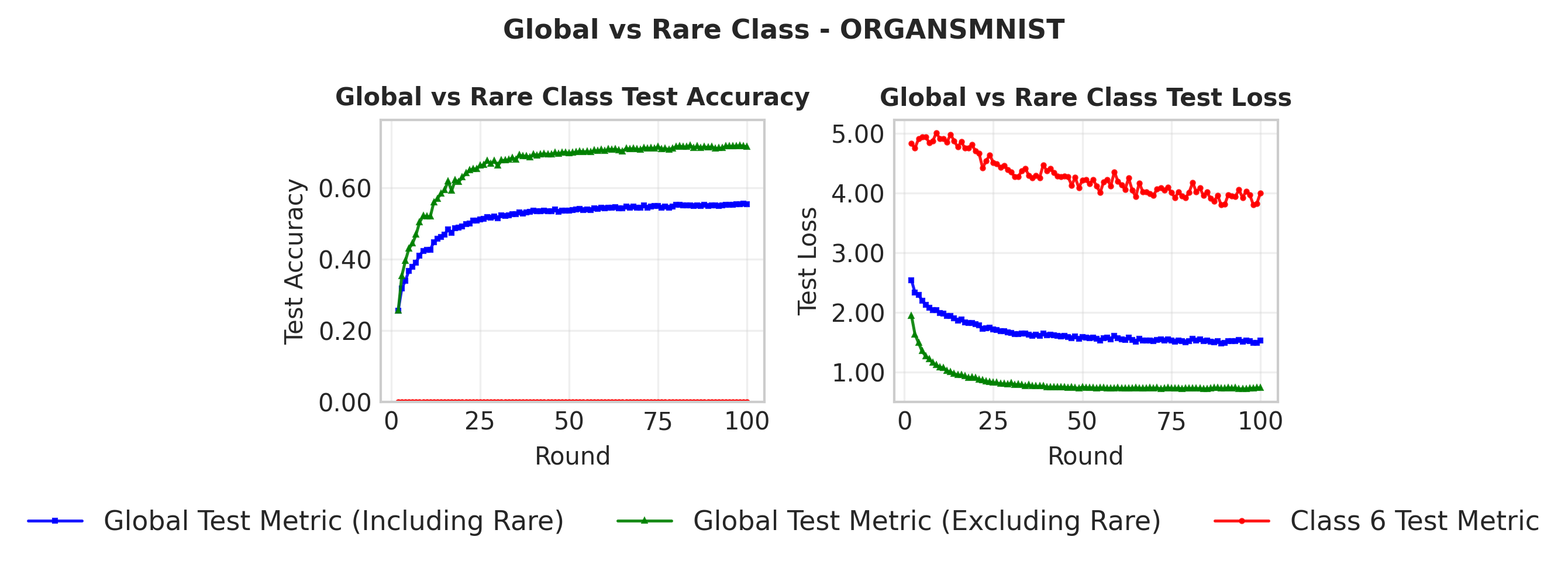}
        \caption{OrganSMNIST: Rare client with $p^{(1)}=0$.}
        \label{figure108}
    \end{subfigure}%
    \hfill
    \begin{subfigure}[t]{0.49\textwidth}
        \centering
        \includegraphics[width=\linewidth]{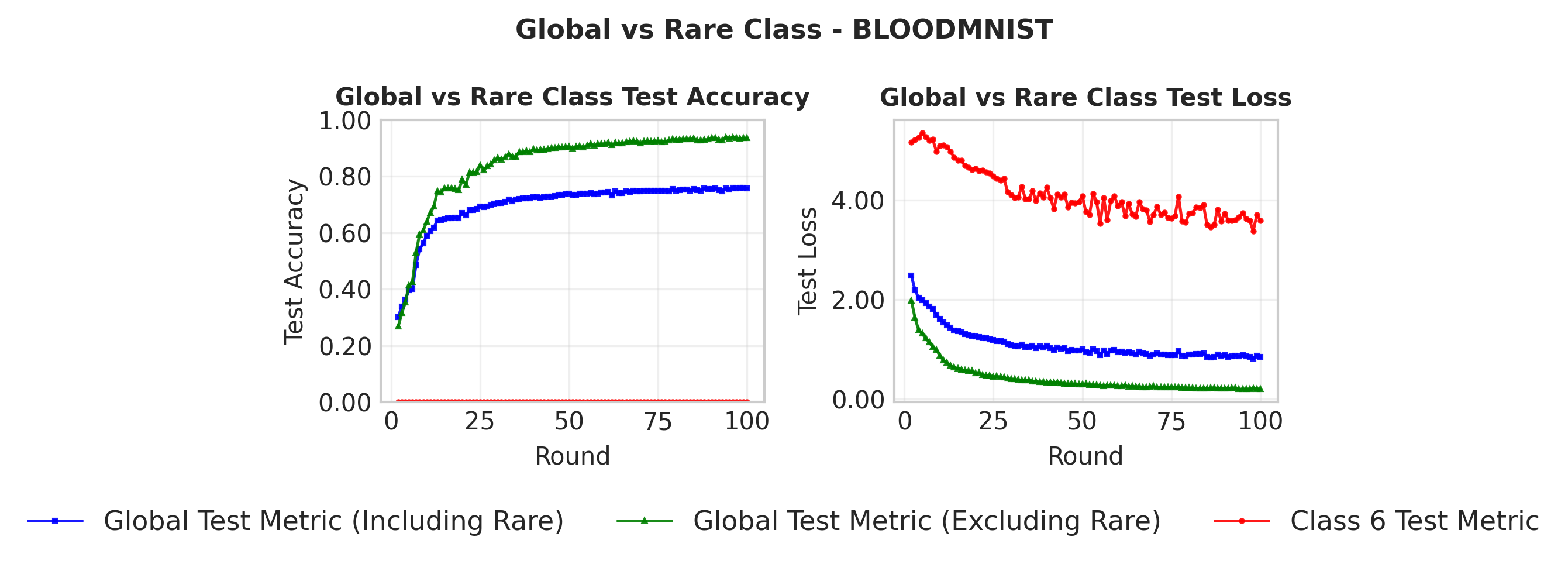}
        \caption{BloodMNIST: Rare client with $p^{(1)}=0$.}
        \label{figure115}
    \end{subfigure}

    % Second row: centered single subfigure
    \par\medskip
    \centering
    \begin{subfigure}[t]{0.49\textwidth} % same width as the first row, but centered
        \centering
        \includegraphics[width=\linewidth]{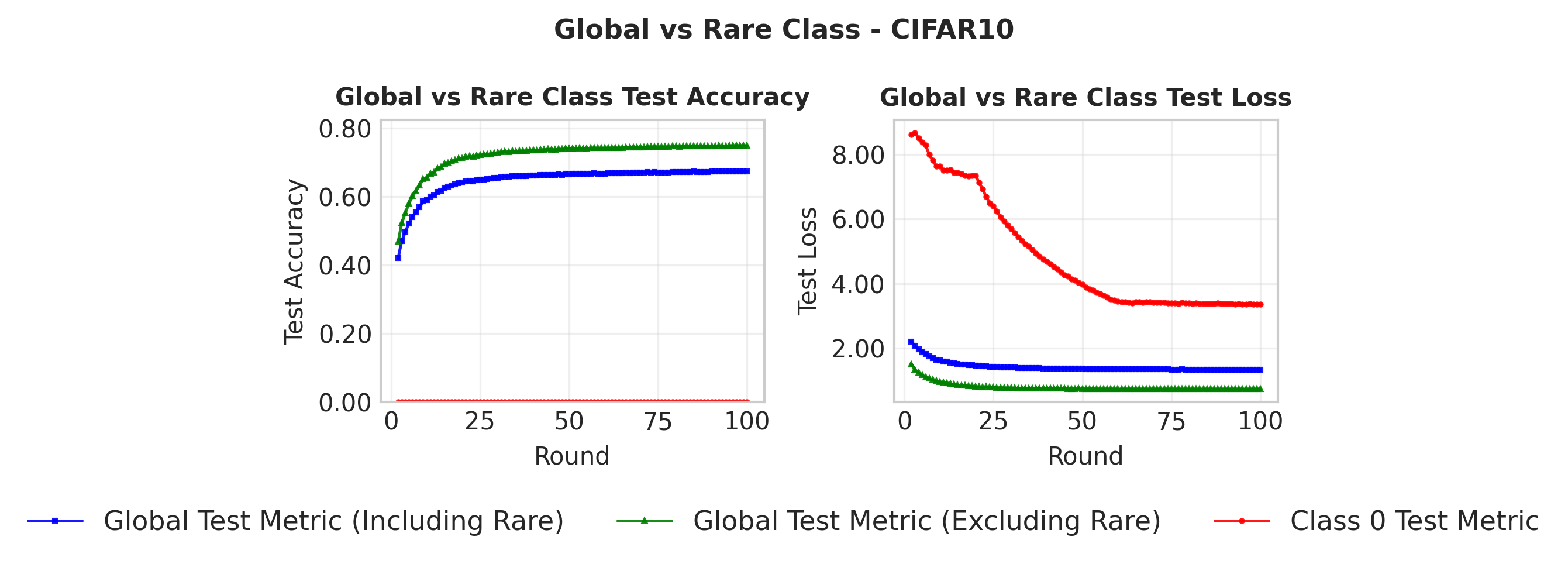}
        \caption{CIFAR-10: Rare client with $p^{(1)}=0$.}
        \label{figure122}
    \end{subfigure}

    \caption{Global test accuracy/loss vs. rare-class performance with $p^{(1)}=0$ (vanishing influence).}
    \label{fig:case3_all}
\end{figure}

\textbf{Unequal Step Sizes with Vanishing influence} appears when a  rare-class client in the previous case uses a faster-decaying schedule  $a_n^{(1)}{=}0.1/n$, making $p^{(1)}=0$. From Figures\ref{figure108},\ref{figure115} and\ref{figure122} it can seen that the rare-client benefits from the global model (as the centralized test metrics improves) without contributing meaningfully to the global model as the rare-class test metrics shows no improvement for all the three datasets. 

{\it It is worth mentioning that if the choices of step sizes are with the client, one or more not wanting to contribute (perhaps due to the exclusivity of his data set) but does not want be seen as not contributing, may resort to using the faster decaying schedule. This can be found out by checking the test losses and accuracy using the held out data of the rare class as shown in Figures\ref{figure108},\ref{figure115} and \ref{figure122}}.

\subsubsection{Comparison with Baseline Algorithms.}
Here the baseline algorithms FedAvg, FedProx and FedNova, with a step size $a^{(i)}_n= 0.1$ for all $i$ is compared with proposed algorithm with  $a^{(i)}_n= 0.1/n^{0.76}$ for all $i$.  It can be observed from Figures\ref{figure112b}
,\ref{figure119b} and\ref{figure126b} for all the three datasets, the baseline algorithms show show persistent oscillations and non-converging $\Delta\bar{w}_{nN}$, while for the proposed method converges to zero. Also, in the case of the training and test metrics persistent oscillations can be seen in the baseline methods while proposed method achieves robust convergence.
\begin{figure}[htb!]
    \centering
        \includegraphics[width=0.75\linewidth]{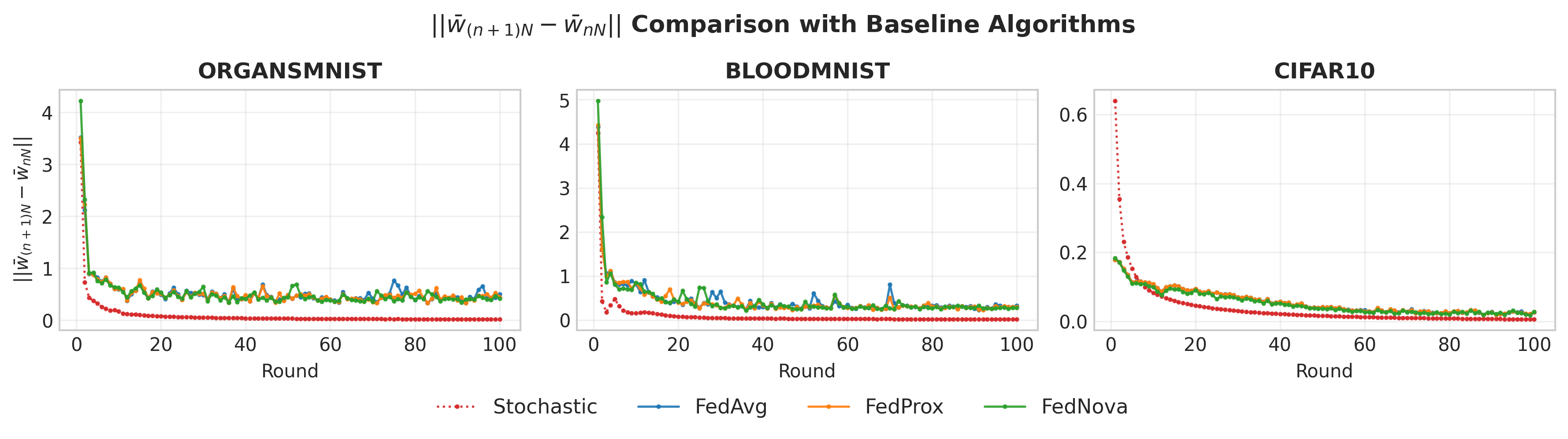}
    \caption{Comparison with Baseline algorithms across datasets. A plot of $\|\bar{w}_{(n+1)N}-\bar{w}_{nN}\|$. Only the proposed method shows convergence.}
    \label{fig:baseline_gradnorm_all}
\end{figure}
\begin{figure}[htb!]
    \begin{subfigure}[t]{0.95\textwidth}
        \includegraphics[width=\linewidth]{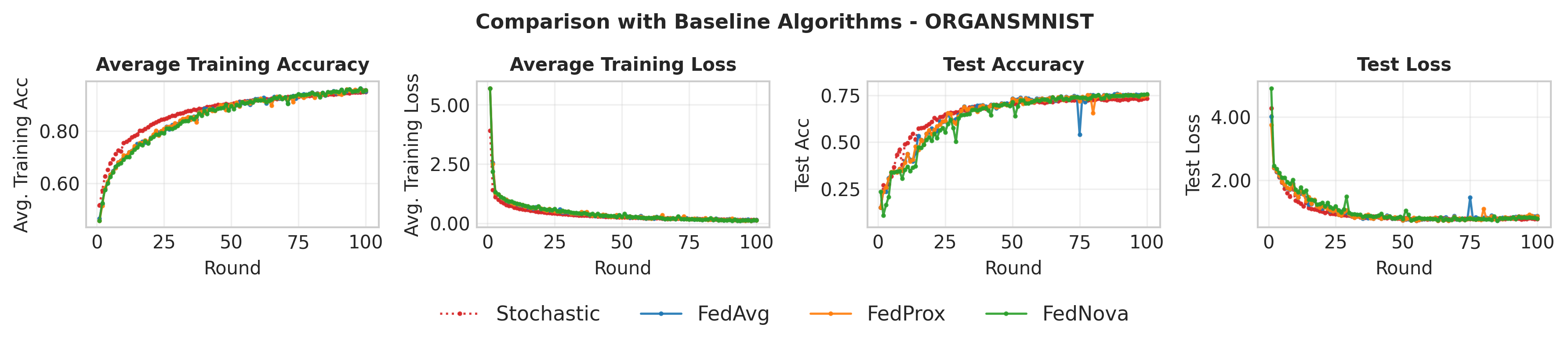}
        \caption{OrganSMNIST}
        \label{figure112b}
    \end{subfigure}
    
    \begin{subfigure}[t]{0.95\textwidth}
        \includegraphics[width=\linewidth]{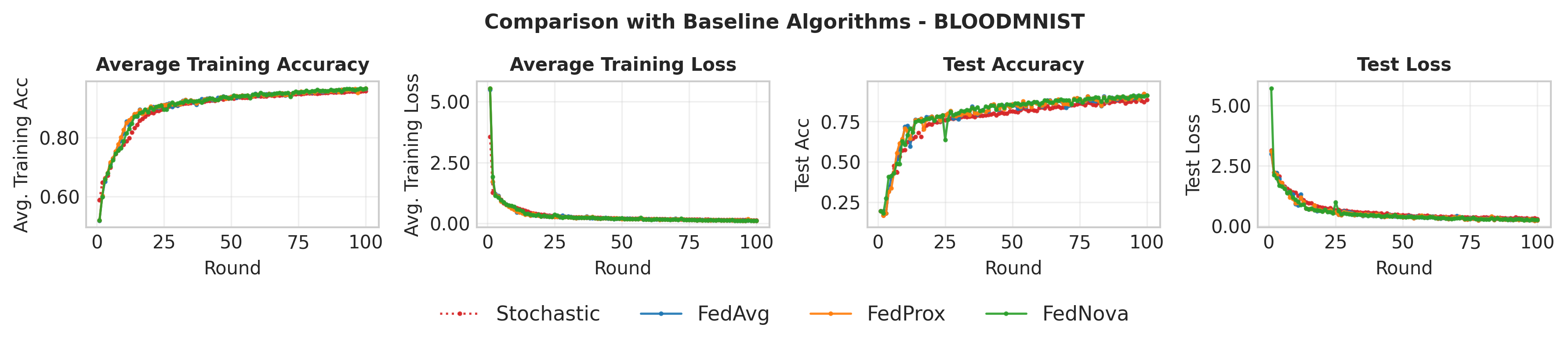}
        \caption{BloodMNIST}
        \label{figure119b}
    \end{subfigure}
    
    \begin{subfigure}[t]{0.95\textwidth}
        \includegraphics[width=\linewidth]{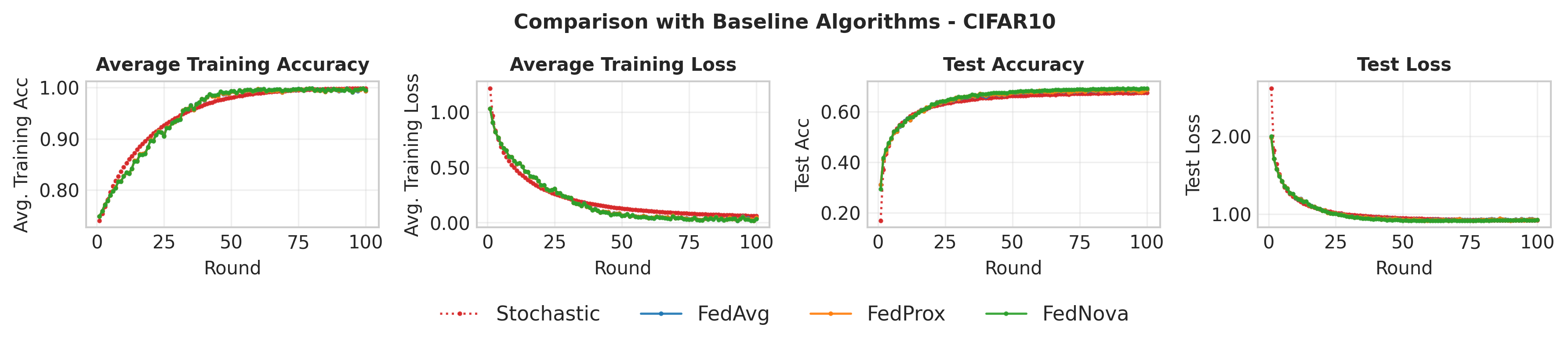}
        \caption{CIFAR-10}
        \label{figure126b}
    \end{subfigure}
    \caption{Baseline comparison across datasets: average training and global test metrics.}
    \label{fig:baseline_metrics_all}
\end{figure}

Figures~~\ref{figure113b}, ~\ref{figure120b} and ~\ref{figure127b} shows the plots of FedAvg, FedProx and FedNova with $a^{(i)}_n=0.1/n^{0.76}$. It is seen that the oscillations are suppressed and $\|\bar{w}_{(n+1)N}-\bar{w}_{nN}\|$ converges to zero. Also, the curves of train and test metrics are smoothened without harming final accuracy.
\begin{figure}[htb!]
    \centering
        \includegraphics[width=0.75\linewidth]{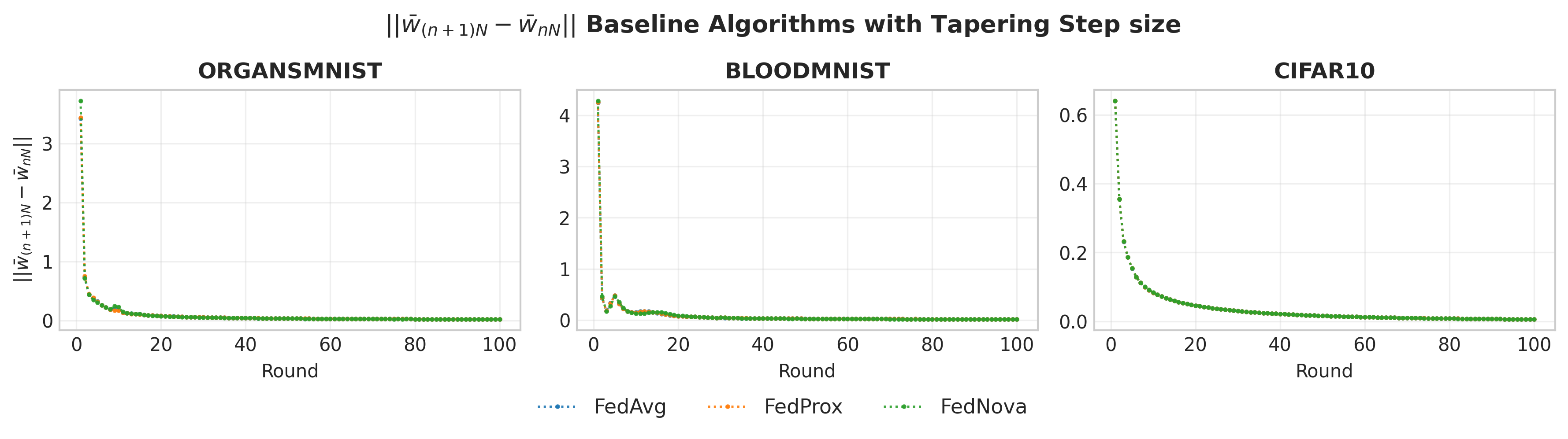}
    \caption{Baseline algorithm with Tapering Step sizes. A plot of $\|\bar{w}_{(n+1)N}-\bar{w}_{nN}\|$ showing convergence for all algorithms. }
    \label{fig:baseline_tapering_gradnorm_all}
\end{figure}
\begin{figure}[htb!]
    \centering
    \begin{subfigure}[t]{0.95\textwidth}
        \includegraphics[width=\linewidth]{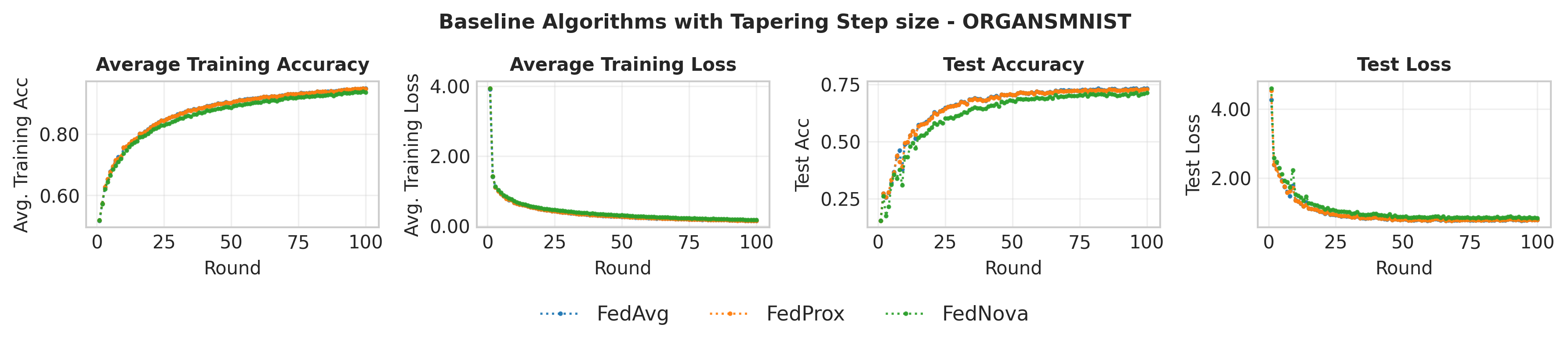}
        \caption{OrganSMNIST}
        \label{figure113b}
    \end{subfigure}
    
    \begin{subfigure}[t]{0.95\textwidth}
        \includegraphics[width=\linewidth]{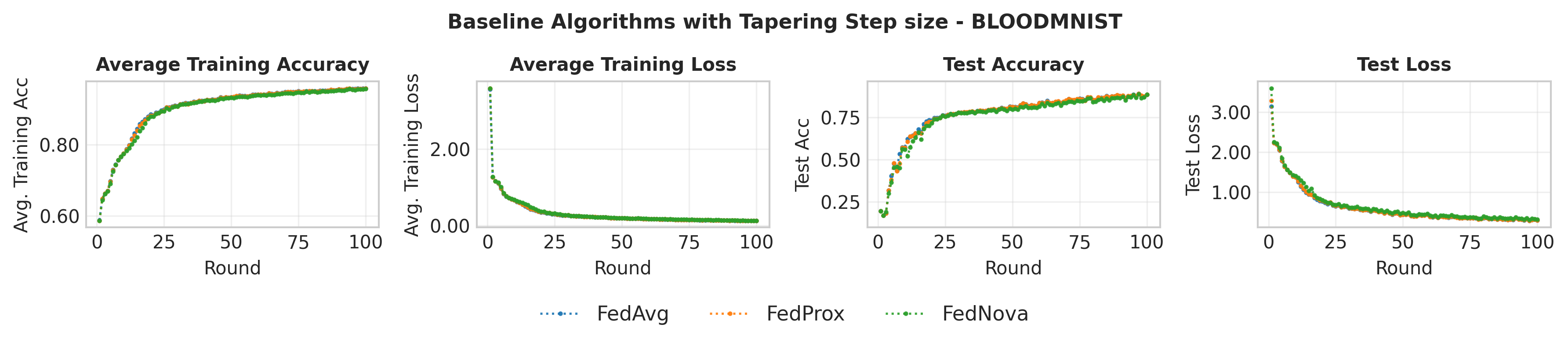}
        \caption{BloodMNIST}
        \label{figure120b}
    \end{subfigure}
    
    \begin{subfigure}[t]{0.95\textwidth}
        \includegraphics[width=\linewidth]{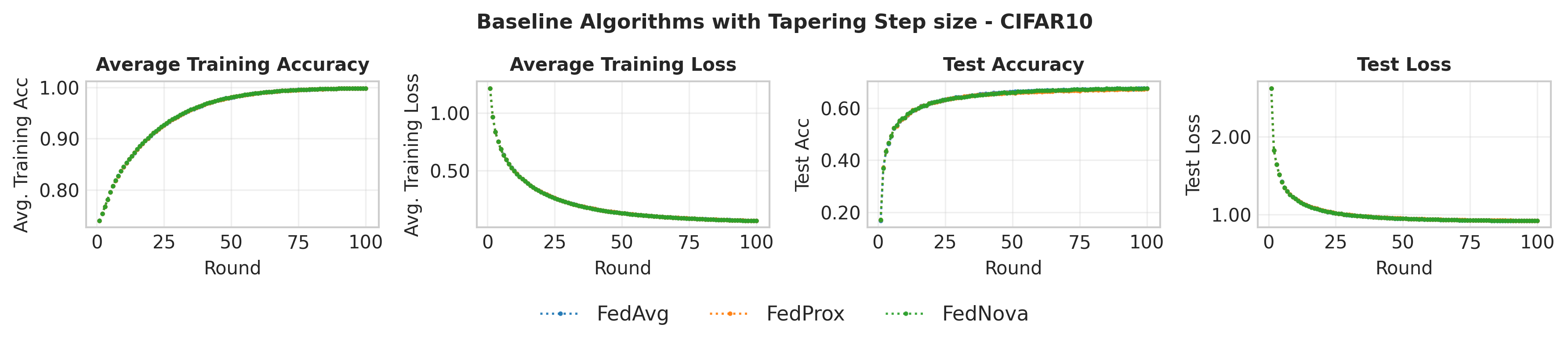}
        \caption{CIFAR-10}
        \label{figure127b}
    \end{subfigure}
    \caption{Baseline algorithms with tapering: average training and global test metrics across datasets.}
    \label{fig:baseline_tapering_metrics_all}
\end{figure}

%\paragraph{Takeaway.}
%In classification, as in regression, \emph{who} influences the global model is dictated by the \emph{relative step-size weights} $p^{(i)}$. Tapering is essential for stable convergence; $\delta$ and $N$ trade speed for smoothness; and controlling $C$ (or $\delta$) can fairly upweight rare-class clients to improve global coverage.

\section{Conclusion}
\label{Conclusion}
This  paper considers an FL setup with each client $i$ training a neural network ${\rm NN}^{(i)}$.  The weights $w^{(i)}_n$ of ${\rm NN}^{(i)}$ are updated using a  mini batch  stochastic gradient descent  with a tapering step size $ a^{(i)}_n$. These  weights  are then aggregated to obtain a global model. It is shown that the aggregated model parameters converge to the equilibrium points of a non-linear ODE with the forcing function being negative of the  weighted sum of the individual gradients (individual NNs cost function gradients). The weights being the limiting ratios $p^{(i)} = \lim_{n \to \infty}a^{(i)}_n/a^{(1)}_n$ assuming $a^{(1)}_n \geq a^{(i)}_n \forall i$.  As the iterates track the ODE  they converge to the equilibrium points $w^*$ of the latter.   These equilibrium points could be global (if the cost function is convex) or a local minima.

As the iterates converge with probability one to the equilibrium points $w^*$ the convergence is more robust when compared with the  baseline algorithms such as FedAvg, FedProx and FedNova.  Unlike the baseline algorithms, which have a constant step size, the proposed frameworks allows for  client specific step size schedules. This offers the flexibility to tune client weights (if needed) to either amplify or reduce their influence on the global model.  These have been numerically validated by applying the propose frame work on  Linear Regression and Image classification problems. The latter using datasets OrganSMNIST, BloodMNIST \& CIFAR-10.

\paragraph{{\bf Future Directions :}}

The client's distribution \(\{\xi^{(i)}_k\}_{k=1}^\infty\) may not be stationary in some datasets (eg. when clients have time series data) thereby violating the Assumption  A4. An example \(\xi^{(i)}_k\) being a Gauss-Markov process. Refer to Appendix F of the Supplementary material for an analysis. More general non-stationary distribution would be of interest for a theoretical investigation. This work assumes global synchronization at \(n=0,N,2N,\ldots\).
A more general setting would allow  asynchronous or time-delayed aggregation for straggler clients. A theoretical investigation into this is possibly the next extension (empirically, it  works). 

%Neonatal seizure detection from EEG is a promising application where privacy concerns and limited hospital-level data pose major challenges. A recent TATPat-based model~\cite{tuncer2024tatpat} demonstrated strong performance but moderate cross-subject generalization, highlighting the difficulty of building models that transfer across patients and clinical centers. The proposed FL framework with client-specific tapering step sizes provides a natural solution where the hospitals can train locally on their EEG data while global aggregation, through tunable weight ratios, help accommodate heterogeneity in datasets. 

Neonatal seizure detection from EEG is a promising application where privacy concerns and siloed (isolated hospital-level data) institutional datasets pose major challenges. A recent TATPat-based model~\cite{tuncer2024tatpat} demonstrated 99.15\% classification accuracy under 10-fold cross-validation but exhibited only 76.37\% accuracy under leave-one-subject-out validation. This highlights the difficulty of building models that generalize across patients and, by extension, clinical centers. The proposed FL framework with client-specific tapering step sizes could provide a natural solution where hospitals can train locally on their EEG data while global aggregation, through tunable weight ratios, helps accommodate heterogeneity in datasets.

%% Use figure environment to create figures
%% Refer following link for more details.
%% https://en.wikibooks.org/wiki/LaTeX/Floats,_Figures_and_Captions

%\appendix

\bibliographystyle{elsarticle-num} 
\bibliography{bibliography}

\end{document}